%% file: main.tex
\documentclass{article}
\usepackage[final,main]{neurips_2026}
\setcitestyle{numbers,square,comma}

\usepackage[utf8]{inputenc} % allow utf-8 input
\usepackage[T1]{fontenc}    % use 8-bit T1 fonts
\usepackage{hyperref}       % hyperlinks
\usepackage{url}            % simple URL typesetting
\usepackage{booktabs}       % professional-quality tables
\usepackage{amsfonts}       % blackboard math symbols
\usepackage{nicefrac}       % compact symbols for 1/2, etc.
\usepackage{microtype}      % microtypography
\usepackage{xcolor}         % colors
\usepackage{float}
\usepackage{graphicx}
\usepackage{svg}
\usepackage{comment}
\usepackage{subcaption}
\usepackage{color,soul}
\usepackage{comment}
\usepackage{amsmath}
\usepackage{amssymb}
\usepackage{mathtools}
\usepackage{amsthm}
\usepackage{algpseudocode}
\usepackage{algorithm}
\usepackage{makecell}
\usepackage{arydshln}

\newcommand{\spikgam}{SpikingGamma}

\definecolor{c0}{HTML}{4C70D6}
\definecolor{c1}{HTML}{EE6677}

\newcommand\codefootnote[1]{%
  \begingroup
  \renewcommand\thefootnote{}% Remove footnote number
  \footnotetext{\hspace{-1em}#1}% Write the text without a marker link
  \endgroup
}

\title{SpikingGamma: Temporally Precise Online SNN Training Through Smoothed Temporal Delays
}

\author{
  Roel Koopman \\
  CWI \\
  \texttt{roel.koopman@cwi.nl}
  \And
  Sebastian Otte \\
  University of Luebeck \\
  \texttt{sebastian.otte@uni-luebeck.de}
  \And
  Sander Bohté \\
  CWI \\
  University of Amsterdam \\
  \texttt{s.m.bohte@cwi.nl}
}

\begin{document}
\maketitle

\begin{abstract}
Neuromorphic hardware implementations of Spiking Neural Networks (SNNs) promise energy-efficient, low-latency AI through sparse, event-driven computation. Yet, training SNNs under fine temporal discretization remains  challenging, hindering low-latency responsiveness and the mapping of software-trained SNNs to efficient hardware. In current approaches, spiking neurons are modeled as self-recurrent units, embedded into recurrent networks to maintain state over time, and trained with BPTT or RTRL variants based on surrogate gradients. These methods scale poorly with temporal resolution, while online approximations exhibit instability for long sequences and tend to fail at capturing temporal patterns. To address these limitations, we develop SpikingGamma, an SNN architecture in which neurons maintain a compact bank of smoothed-delay states and communicate through sigma-delta spike-coding. We show that in feedforward networks, \spikgam{} supports direct online error backpropagation through the continuous reconstruction signal, avoiding surrogate gradients through the spike discontinuity. This enables stable learning of temporal patterns with minimal spiking and scales feedforward SNNs to complex tasks and benchmarks with competitive accuracy, all while remaining robust to the temporal resolution of the model. Our approach offers both an alternative to recurrent SNNs trained with surrogate gradients, and a novel route for mapping SNNs to neuromorphic hardware.
\end{abstract}

\codefootnote{
Code is available at: \url{https://github.com/RoelMK/SpikingGamma}
}

\section{Introduction}

Inspired by the brain, Spiking Neural Networks (SNNs) \cite{maass1997networks} hold promise for energy-efficient AI models \cite{li2023efficient, dampfhoffer2022snns} as they use sparse and discrete spikes for communication and enable event-driven, asynchronous information processing.  Spiking neurons can thus respond only when needed, achieving low-latency responsiveness while lowering energy consumption, especially when tasks exhibit extended temporal dynamics.  
Consequently, SNNs are well-suited for real-world tasks such as autonomous driving \cite{martinez2024spiking}, auditory signal processing \cite{baek2024snn}, and drone control \cite{hagenaars2020evolved}. Treating spiking neurons as self-recurrent neural units and 
using a so-called Surrogate Gradient (SG) to backpropagate errors through the discontinuity of the spiking process \cite{neftci2019surrogate}, current SNNs now demonstrate competitive performance with classical neural networks for small scale tasks \cite{hammouamri2024learning,eshraghian2023training}. 

Yet, since current SNNs are typically treated as time-stepped Recurrent Neural Networks (RNNs), they inherit the training drawbacks of recurrent models: vanishing and exploding gradients, and the need to preserve or approximate past influences on current state. Exact methods such as Backpropagation Through Time (BPTT) and Real-Time Recurrent Learning (RTRL) \cite{zenke2021remarkable} incur prohibitive memory or timestep complexity, while online approximations \cite{kaiser2020synaptic,wang2024brainscale} often degrade when scaling in sequence length or network size. Surrogate gradients further compound this issue and can limit attainable sparsity, as excessive sparsity amplifies gradient vanishing and can collapse training \cite{zenke2021remarkable}.

This scaling problem is especially restrictive for hardware-compatible SNNs. Low-latency operation and faithful deployment to neuromorphic substrates require training over many small timesteps,  so model dynamics must closely match physical device dynamics  \cite{cramer2022surrogate,ko2024snnsim,koopman2025exploring}. Fine-grained temporal simulation therefore directly conflicts with the dominant recurrent SNN training paradigm, including, as we show, online BPTT/RTRL approximations such as FPTT \cite{kag2021training} and ES-D-RTRL \cite{wang2024brainscale}.

Here, we introduce the \spikgam{} model, which enables online training at arbitrarily fine temporal resolution. Building on causal-delay models including the Gamma-model \cite{de1992gamma}, TKRNN \cite{sutskever2010temporal}, and Fractionally Predictive SNNs \cite{bohte2011error,rombouts2010fractionally}, \spikgam{} stores temporal history in recursively computed smoothed-delay kernels rather than recurrent state. Unlike discrete-delay SNNs \cite{hammouamri2024learning}, whose memory requirements grow linearly with temporal resolution, this representation can be evaluated at finer temporal discretizations without introducing a corresponding memory cost.

In the \spikgam{} model, spiking neurons use adaptive recursive memory to construct delayed representations of past inputs, while sigma-delta spike-coding \cite{yoon2016lif,rombouts2010fractionally} encodes the rectified internal state into a spike train that downstream neurons can reconstruct, analogous to pulsed sigma-delta coding in electrical circuits \cite{yoon2016lif}. This combination removes self-recurrency and allows feedforward SNNs to be trained directly by error backpropagation rather than BPTT, without surrogate gradients through the spiking discontinuity. The resulting internal delays allow \spikgam{} neurons to detect co-occurrences of temporally disjoint features without externally maintained memory, such as persistent activity. \spikgam{} thus enables a causal feedforward model where the state summarizing past presynaptic events replaces the unrolled recurrent state for temporal credit assignment while retaining spike-based communication.

Here, we show how feedforward \spikgam{} SNNs can compete with state-of-the-art recurrent SNN approaches. We show that \spikgam{} SNNs can learn sparse and precise temporal coding over long timespans, while being insensitive to the temporal resolution of the simulated dynamics. We demonstrate this for both simple delay tasks, an echo-location task, and standard neuromorphic benchmarks such as DVS Gesture, SHD and SSC, where we show superior accuracy compared to current online methods with scalable temporal resolution.  
Our approach thus presents an alternative to current recurrent SNNs based on Surrogate Gradient training, where \spikgam{} SNNs can be trained and computed online and at arbitrarily high temporal resolution. 

\section{Related work}
Deep SNNs have been trained via ANN-to-SNN conversion, spiking variants of backpropagation, and biologically motivated plasticity rules such as STDP, each with distinct trade-offs in accuracy and hardware efficiency \cite{pfeiffer2018deep}. Where surrogate gradients have enabled training of SNNs with standard recurrent deep learning approaches, several alternatives to BPTT and RTRL have been explored to improve the training efficiency of SNNs. ANN-to-SNN conversion, while straightforward and scalable, fails to exploit the temporal dynamics and event-driven sparsity of SNNs, and most existing methods remain constrained to inefficient rate-based coding schemes \cite{zhou2024direct}. Various approximations of RTRL and BPTT have been proposed that maintain temporal traces for credit assignment or regularize gradients through time rather than computing exact gradients, such as DECOLLE \cite{kaiser2020synaptic}, OSTL \cite{bohnstingl2022online}, OTPE \cite{summe2024estimating}, FPTT \cite{yin2023fptt}, OTTT \cite{xiao2022online}, and ES-D-RTRL \cite{wang2024brainscale}. These methods have shown limited success on tasks demanding complex temporal reasoning. Biologically motivated work approximates backpropagation through space and time online, using only local signals \cite{ellenberger2025backpropagation}. Another line of work seeks to exploit the inherent sparsity of SNN activity to reduce the memory footprint of BPTT, such as EventProp \cite{meszaros2025efficient}, though its scalability to large models and datasets remains unexplored.

Our work is inspired by a different idea: maintaining a compressed representation of history within the forward pass to enable temporal learning without requiring BPTT. Related ideas have been explored in ANNs. For example, in the TKRNN \cite{sutskever2010temporal} all neurons are linked through all timesteps using a learned weight matrix, akin to delays, WaveNet \cite{van2016wavenet} uses dilated causal convolutions to model long-range temporal dependencies in a feedforward architecture, performing comparably to LSTM-based RNNs, and Legendre Memory Units \cite{voelker2019legendre} compress a sliding input window into a fixed-size continuous-time state, but inside a recurrent cell trained with BPTT.
The Gamma Model \cite{de1992gamma} similarly maintains adaptive internal memory representing past inputs, functioning like a form of smoothed temporal convolution. Notably, so-called time-cells have been identified in the brain with similar delayed response properties \cite{liu2019neural}, and Gamma-models have been proposed to account for internal dynamics in biological neurons such as spike-rate adaptation \cite{drew2006models}. Smoothed-delay representations have also been used to model temporally disjunct cue–reward associations in neuroscience, but not scaled to multilayer SNNs or complex temporal benchmarks \cite{ludvig2008stimulus}. A related approach was applied to SNNs to avoid BPTT \cite{bohte2011error}, that delay-structure however is incompatible with scalable learning as it cannot be computed in a recurrent manner -- the concepts introduced in \cite{de1992gamma} and \cite{bohte2011error} however form the basis for our work. 

\section{Methods}

\subsection{The neuron model}

\begin{figure}[!htb]
    \centering
    
    \begin{subfigure}[t]{0.53\linewidth}
        \centering
        \includegraphics[width=\linewidth]{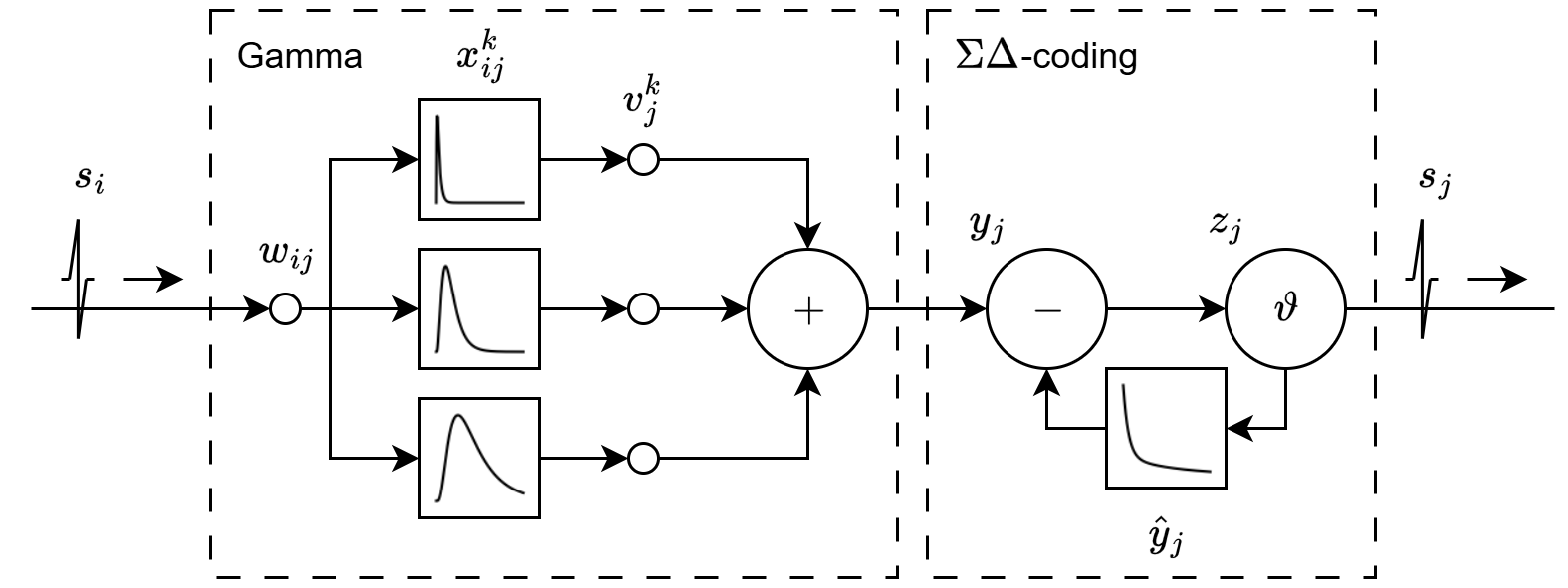}
        \caption{}
        \label{fig:neuron_model_overview}
    \end{subfigure}
    \hfill
    \begin{subfigure}[t]{0.43\linewidth}
        \centering
        \includegraphics[width=\linewidth]{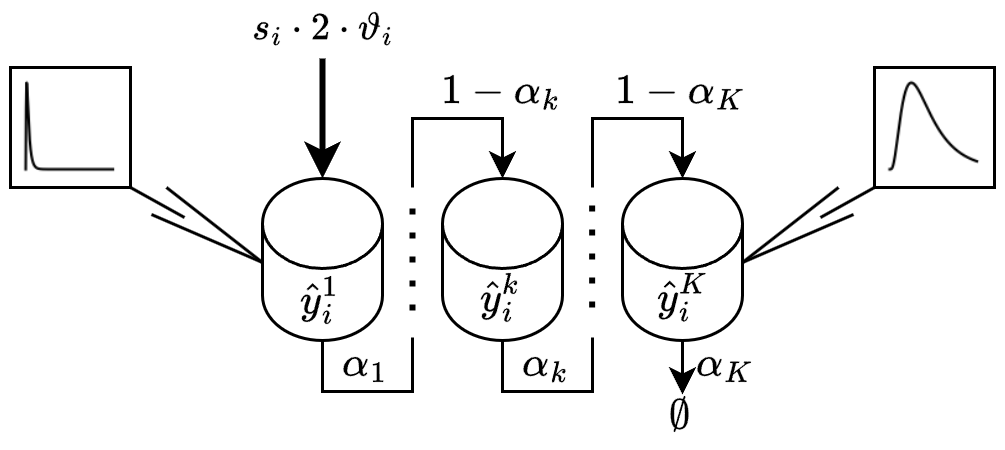}
        \caption{}
        \label{fig:buckets}
    \end{subfigure}
    
    \caption{
    (a) Neural processing model. Incoming spikes generate weighted synaptic currents across multiple timescales. The neuron weights and sums these responses into a continuous signal, which is converted back to spikes via $\Sigma\Delta$-coding so downstream synapses can reconstruct it.
(b) Temporal kernel computation as a cascade of leaky “buckets” draining into one another at rates $\alpha_k$, with each bucket representing one kernel.
    }
    
    \label{fig:combined_neuron_model}
\end{figure}

In the SpikingGamma model, we base the spiking neuron model on a Linear-Non-Linear Sigma Delta (LNL-SD) temporal filtering model \cite{rombouts2010fractionally}, where a neuron $j$ computes an internal signal $y_j$ as the sum of the weighted and filtered inputs into the neuron. To implement sigma-delta spike-coding, the neuron additionally tracks a signal $\hat{y}_j$ as the sum of spike-triggered refractory responses \cite{gerstner2002spiking}: as in sigma-delta coding \cite{yoon2016lif}, this sum of refractory responses approximates the rectified $y_j$ by emitting a spike and adding a refractory response whenever the positive approximation error between $\hat{y}_j$ and $y_j$ exceeds a threshold $\vartheta$ \cite{zambrano2019sparse}. The approximation of $y_j$ is thus encoded by the spike-train determined by those threshold exceedances as downstream neurons can reconstruct the signal $\hat{y}_j$ at their input -- the neural processing model is illustrated in Figure \ref{fig:neuron_model_overview}.
Mathematically, the signal $y_j$ is computed as:
\begin{equation}
y_j(t) = \text{ReLU}(x_j(t))
\end{equation}
with $x_j$ the unrectified neuron signal. This is computed by filtering the synaptic input signals $x_{ij}^k$ from neuron~$i$ to neuron~$j$ for $k < K$ with each synapse having $K$ temporal kernels. The filters can be instantiated as per-neuron or per-synapse. For per-neuron this is described as:
\begin{equation}\label{eq:xj}
x_{j}(t) = \sum_{k} \sum_i x_{ij}^k(t) \cdot v_j^k
\end{equation}
where $v_j^k$ is a parameter that weights the value of kernel~$k$ for neuron~$j$. We refer to this as the ``bucket weight''.
The synaptic signal $x_{ij}^k$ is then computed as: 
\begin{equation}\label{eq:xjk}
    x_{ij}^k(t) = \hat{y}_{i}^k(t)\cdot w_{ij}
\end{equation}
where $w_{ij}$ is the synaptic weight. The term $\hat{y}_{i}^k(t)$ denotes the contribution of temporal kernel $k$ at time $t$ to the estimate of the signal of neuron $i$, as encoded by the output spikes $t_i$ from neuron $i$ to neuron $j$:
\begin{equation}
\label{eq:yhat}
    \hat{y}_{i}^k(t) = \sum_{t_i \leq t}\kappa^k(t-t_i)
\end{equation}
where $t_i$ denotes spike-times of neuron~$i$ up to time $t$, and $\kappa^k$
denoting a set of temporal kernel functions implementing different delays. Amplitude modulation is ignored here.

Following the Gamma-model \cite{de1992gamma,drew2006models} the kernels $\kappa^k$ are computed as a series of ``buckets'' that spill over into each other. This is illustrated in Figure \ref{fig:buckets}, and described as: 
\begin{equation}\label{eq:bucket_dynamics}
\hat{y}_i^k(t)=
\begin{cases}
\hat{y}_i^k(t-1)\alpha_k
+ s_i(t)\,2\,\vartheta_i(t)
& k=0, \\[4pt]
\hat{y}_i^k(t-1)\alpha_k
+ \hat{y}_i^{k-1}(t-1)(1-\alpha_k)
& k>0.
\end{cases}
\end{equation}
with $\alpha_k$ specifying the transfer rate between buckets $k$ and $k-1$, $\vartheta_i(t)$ denoting the threshold function of the spiking neuron, as defined in Section \ref{sec:adaptive_thresholding}, and $s_i(t) = 1$ if $t \in \{t_i\}$ for an output spike-train $\{t_i\}$, else $0$. 
The factor 2 ensures that $\hat{y}$ matches the magnitude of $y$ instead of being biased low \cite{zambrano2019sparse}.
As this is a set of linearly coupled differential equations, each $\hat{y}_i^k$ can be reformulated as Eq. \eqref{eq:yhat} 
with each $\kappa^k$ not relying on any other kernel: each $\kappa^k$, and thus each bucket $\hat{y}_{i}^k$ can be modeled in a feedforward fashion without recurrency -- that is, knowing the spike times of the inputs $t_i < t$, one can exactly compute the value of $\hat{y}_{i}^k(t)$ for any~$t$ \cite{bohte2011error}.  

\begin{figure*}[!tb]
    \centering
    \includegraphics[width=\textwidth] {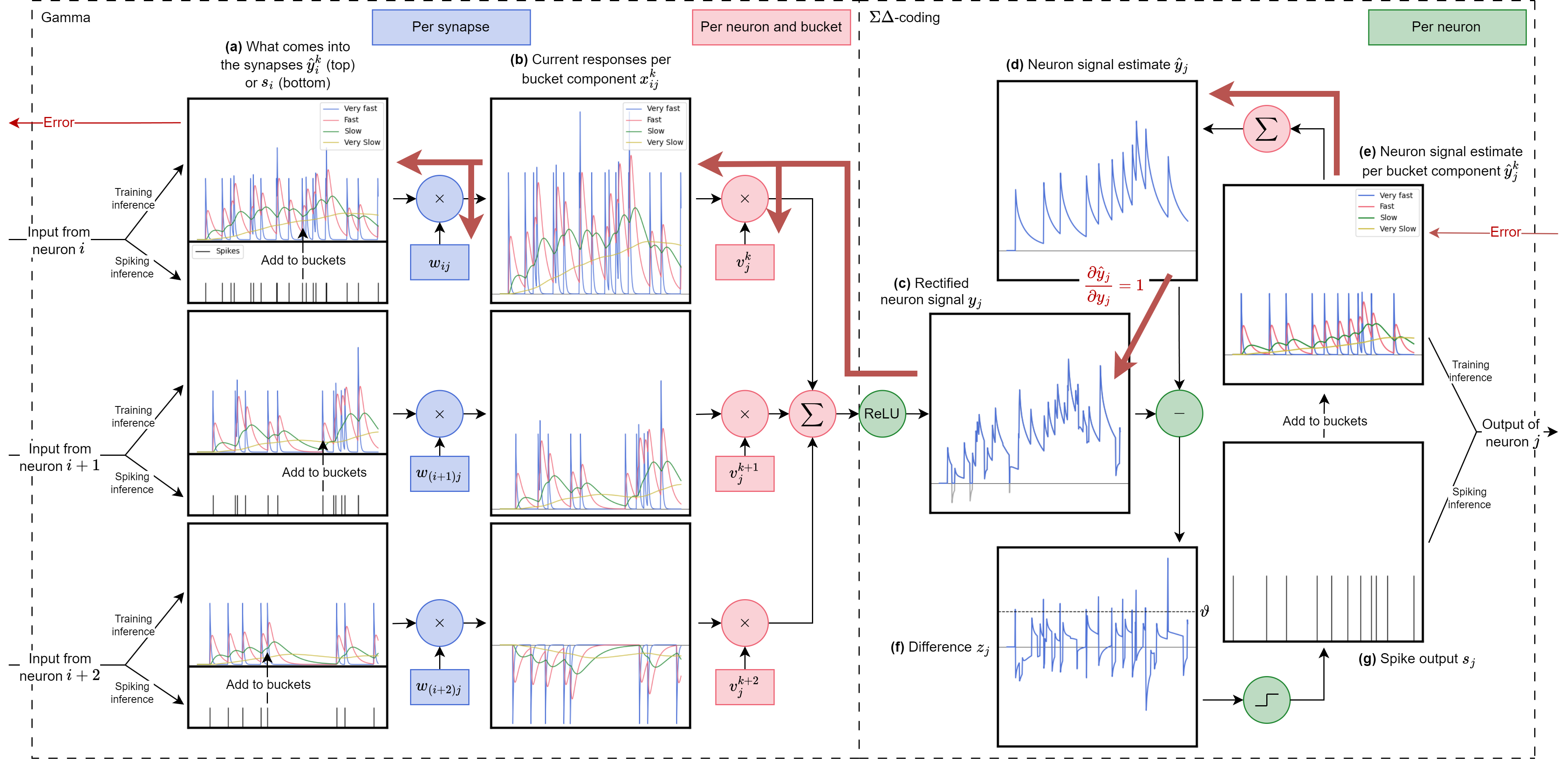} 
    \caption{
    Signal evolution and error propagation in the neuron model. 
\textbf{(a)} Input synapses receive presynaptic signal estimates $\hat{y}_i^k$ during training, or reconstruct them from incoming spikes $s_i$ during spiking inference via Eq.~\eqref{eq:bucket_dynamics}. 
\textbf{(b)} Synaptic weights $w_{ij}$ scale each input into per-bucket current responses. The negative weight $w_{(i+2)j}$ inverts the responses of input $i+2$. 
\textbf{(c)} Bucket weights $v_j^k$ (or $v_{ij}^k$ for per-synapse weighting) combine these responses, which are summed and rectified to form $y_j$. The horizontal line marks zero, and gray shows the negative values removed by rectification.
\textbf{(d--g)} Sigma-delta coding maintains an estimate $\hat{y}_j$ in the same kernel basis, emits a spike $s_j$ when the mismatch $z_j = y_j - \hat{y}_j$ exceeds threshold, and adds this correction back into the estimate. 
Because $\hat{y}_j^k$ and $\hat{y}_i^k$ share the same representation, $\hat{y}_j^k$ can be passed directly to downstream synapses during training. 
Red arrows denote the training error path: gradients flow from the signal estimate to the neuron signal, inputs, and parameters, bypassing spikes and avoiding surrogate gradients.    
    }
    \label{fig:structure_neuron_model_with_signals}
\end{figure*}

Spike thresholding implements sigma-delta spike-coding \cite{yoon2016lif,rombouts2010fractionally} and is done by first computing a variable $z_j(t)$ that is similar to the membrane potential by subtracting an approximation of the signal $\hat{y}_j$, which is what will be encoded by the to be emitted spikes, from the actual internal signal $y_j$:
\begin{equation}
z_j(t) = y_j(t) - \hat{y}_j(t-1), \quad \text{with} \quad \hat{y}_j(t) = \sum_k \hat{y}_j^k(t)
\end{equation}
where $\hat{y}_j^k$ is defined per Eq. \eqref{eq:bucket_dynamics} for the output of neuron $j$ itself. When the difference exceeds a threshold, the signal approximation $\hat{y}_j(t)$ is then updated by emitting a new spike: 
\begin{equation}
s_j(t) = z_j(t) > \vartheta_j(t), \quad t_j = t \;\text{if } s_j(t) = 1
\end{equation}
with $\{t_j\}$ denoting the output spike-train. Figure \ref{fig:structure_neuron_model_with_signals} visualizes the complete forward path together with example signal values at each stage. Pseudocode is provided in Appendix~\ref{appendix:pseudocode}.

Thus formulated, both the signal approximation at the output of the neuron and the signal received at the downstream synapses are the same, making it possible to directly use $\hat{y}_j^k(t)$ for computations at those synapses (i.e., for Eq. \eqref{eq:xjk}). To filter per synapse, the only change needed to the model above is to use individual per-synapse bucket weights $v_{ij}^k$ instead of $v_{j}^k$. 
Notably, SpikingGamma can be interpreted as a nonlinear SSM with linear latent dynamics and dual readouts (Appendix \ref{math:ssm}).

\subsection{Error-backpropagation}\label{sec:error_backprop}

For classification, given some desired output label $y_\text{true}(t)$ and the actual neuron signal of the output neuron $y_\text{out}(t)$, we compute a loss such as the Cross-Entropy (CE) loss $L(t) = L^\text{CE}(y_\text{out}(t), y_\text{true}(t))$, evaluated independently at each discrete timestep, since temporal credit assignment is handled explicitly via bucketed histories rather than recurrent state (see Figure \ref{fig:time_credit}). If the task requires a precise spike timing, we instead use a Mean Squared Error (MSE) loss between the expected $\hat{y}_\text{true}(t)$ that would follow from a spike in the correct class at the right time, and actual neuron output $\hat{y}_\text{out}(t)$. Then, $L(t) = L^\text{MSE}(\hat{y}_\text{out}(t), \hat{y}_\text{true}(t))$.

As illustrated in Figure \ref{fig:structure_neuron_model_with_signals}, for each discrete timestep $t$, we use the variables at that timestep to compute the loss with respect to the synaptic weights $w_{ij}$ and the bucket weights $v_j^k$ (or $v_{ij}^k$ in the case of per-synapse filtering) (we omit the explicit time index $(t)$):
\vspace{-0.25cm}
\begin{figure}[H]
    \centering
    \begin{minipage}[t]{0.35\linewidth}
        \vspace{0pt}
        \begin{equation}\label{eq:dL_dwij}
        \begin{aligned}
        \frac{\partial L}{\partial w_{ij}} &=
        \frac{\partial L}{\partial \hat{y}_j}
        \frac{\partial \hat{y}_j}{\partial y_j}
        \frac{\partial y_j}{\partial x_j} \\
        &\quad \times \sum_k \frac{\partial x_j}{\partial x_{ij}^k}
        \frac{\partial x_{ij}^k}{\partial w_{ij}}
        \end{aligned}
        \end{equation}
        \begin{equation}
            \frac{\partial L}{\partial v_{j}^k} =
            \frac{\partial L}{\partial \hat{y}_j}
            \frac{\partial \hat{y}_j}{\partial y_j}
            \frac{\partial y_j}{\partial x_j}
            \frac{\partial x_j}{\partial v_j^k}
        \end{equation}
    \end{minipage}
    \hfill
    \begin{minipage}[t]{0.61\linewidth}
        \vspace{0pt}
        \centering
        \includegraphics[width=0.8\linewidth]{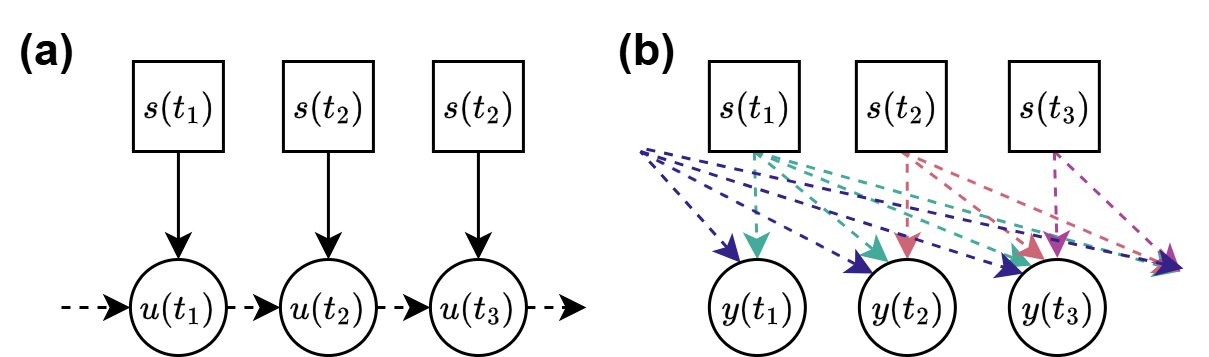}
        \caption{
        Transition of neuron states over time in recurrent architectures (a) versus \spikgam{} (b). Recurrent models rely on internal states (e.g., membrane potential $u$) and require BPTT for training, while \spikgam{} does not as it has access to the entire history via buckets for each timestep.
        } 
        \label{fig:time_credit}
    \end{minipage}
\end{figure}
\vspace{-0.25cm}

where $\frac{\partial L}{\partial \hat{y}_j}$ follows directly from the definition of the loss function. We avoid differentiating through the binary spike-generation function by propagating the error through the continuous reconstruction path, using the identity estimator $\partial \hat{y}_j / \partial y_j \approx 1$, justified when $\hat{y}_j \approx y_j$ through sigma-delta coding (see Appendix~\ref{appendix:estimator_analysis} for an empirical analysis of this approximation).

The other terms follow directly from their forward pass, namely: $\frac{\partial y_j}{\partial x_j} = \text{ReLU}'(x_j)$, $\frac{\partial x_j}{\partial x_{ij}^k} = v_j^k$, and finally $\frac{\partial x_{ij}^k}{\partial w_{ij}} = \hat{y}_i^k$ and $\frac{\partial x_j}{\partial v_j^k} = \sum_i x_{ij}^k$ (again, replace $v_j^k$ with $v_{ij}^k$ for per synapse filtering). 
To traverse the computational tree over a layer, we use $\frac{\partial x_{ij}^k}{\partial \hat{y}_i^k} = w_{ij}$. Since neuron $i$ produces $K$ components $\hat{y}_i^k$, their gradients are aggregated as
$\frac{\partial L}{\partial y_i} = \sum_k \frac{\partial L}{\partial \hat{y}_i^k}$
to continue backpropagation.

\subsection{Initialization, regularization and constraints}
Before training, parameters are initialized. During training, we apply regularization and add constraints to prevent overfitting, improve model generalization, and increase sparsity. We use standard methods as summarized below, and additionally describe our \emph{Adaptive Thresholding} and \emph{Bucket-transfer Rate Initialization} in detail.

\textbf{Dropout on neuron signal}. Randomly zeros out the neuron signal $y$ with a given probability during training. The decision to drop is independent according to a Bernoulli distribution. Applied to every layer except the output layer.

\textbf{Layer normalization}. Normalizes the neuron signal before rectification ($x$) based on layer statistics \cite{ba2016layer}: $x_\text{norm} = \frac{x-\text{E}[x]}{\sqrt{\text{Var}[x]+\epsilon}}\cdot \gamma + \beta$, with $\gamma$ the gain, $\beta$ the bias (both trainable), and $\text{E}[x]$ and $\text{Var}[x]$ respectively the mean and variance of the neuron signal over all neurons in the layer. Group normalization can be used as an alternative when specified (see Appendix~\ref{appendix:group_norm} for details).

\textbf{Gain loss}. Penalizes the magnitude of the gain term $\gamma_l$ of the normalization layers: $L_\text{gain} = G \cdot \sum_l \gamma_l^2$, where $G$ is a constant that affects the relative importance of the gain loss term, and $l$ is the normalization layer index. This shrinks the signal, in turn reducing the number of spikes.

\textbf{Weight initialization}. Synaptic weights and biases on fully-connected and convolutional layers are sampled from  $\mathcal{U}(-\sqrt{k}, \sqrt{k})$ with $k=\frac{1}{\text{input features}}$.
The bucket weights are sampled from $\mathcal{N}(0,0.1)$. Layer normalization gains and biases are initialized to 1 and 0, respectively.

\paragraph{Adaptive thresholding.}\label{sec:adaptive_thresholding}
In the sigma-delta spike-coding model, whenever the difference between $y$ and $\hat{y}$ exceeds a fixed threshold, a constant input  $s$ (pulse) is added to $\hat{y}$. As $y$ increases, the leak from the buckets (which grows relative to $\hat{y}$) eventually balances this constant input. Beyond that point, if $y$ reaches an even higher value, the leak will prevent $\hat{y}$ from approaching $y$. Consequently, the relation $\frac{\partial \hat{y}}{\partial y} = 1$ no longer holds, which can disrupt training.

To overcome this, we use an adaptive thresholding mechanism \cite{zambrano2019sparse} that increases the threshold, and therefore the amount added to the buckets, in proportion to $\hat{y}(t-1)$:
\begin{equation}
\vartheta(t) = \vartheta_0 + \hat{y}(t-1) \cdot m_f
\end{equation}
where $\vartheta_0$ is the minimum threshold, and $m_f$ is a constant scaling factor. In our experiments, we set $m_f = \vartheta_0$. With this adaptive thresholding, $\hat{y}$ can approximately track any positive value of $y$ (Figure \ref{fig:adaptive_thresholding_range_increase}, Appendix). With adaptive thresholds, binary spikes suffice only if each receiving synapse replicates the sender's threshold by tracking $\hat{y}$ per synapse rather than per neuron, at the cost of extra per-synapse state and computation. Otherwise, spikes must carry the threshold value as a graded payload \cite{zambrano2019sparse}, which requires hardware support for graded spikes.

\paragraph{Bucket transfer rate initialization.}\label{sec:param_bucket_transfer_rate}
Following \cite{bohte2011error}, we ensure that the combined sum of buckets follows a power-law-like curve to facilitate learning multi-timescale temporal filtering. To achieve this, we generate linearly separated values $l_k \in \text{linspace}(L_\text{start}, L_\text{end}, K)$, with $L_\text{start}\in (0,1)$ denoting the starting value, $L_\text{end}\in (0,1)$ the final value (both included in the generated value range), and $K$ the number of values, one for each bucket. This is then used to compute the transfer rates $\alpha_k = (l_k)^F$,
with $F \in (0,1)$ the transfer rate factor. Combined with $L_\text{start}$, $L_\text{end}$, and $K$ this determines the shape of the curve. For all experiments, $L_\text{start} = 0.001$ and $L_\text{end} = 0.86$. In Table \ref{tab:results_alternative_kernels} (Appendix), the neuron response for different values of $F$ are shown.

\section{Experiments and results}
We demonstrate how deep feedforward \spikgam{} SNNs can (1) learn to detect and respond to precisely timed spikes, with minimal spiking, (2) achieve accuracy in temporally sensitive benchmarks competitive with and exceeding current online learning approaches for SNNs, and, (3) scale without loss of accuracy to fine-temporal resolutions which are infeasible with exact recurrent learning and where current approximate online learning approaches fail. 

\subsection{Learning exact timings through delayed responses}\label{sec:learning_exact_timings}
We demonstrate the ability of \spikgam{} to efficiently learn to detect fine temporal structure in two examples involving temporally structured computations. 

\paragraph{Learning delays.} In the first task, a single input spike needs to be propagated with a precise delay through a hidden neuron to a single output neuron. This task allows us to compare BPTT learning with BP learning in the feedforward \spikgam{} model. 

Specifically, an input is provided at $t=0$, and the network is trained to produce an output at $t=150$. Each of the two trainable layers in the network contains a single neuron, with only the bucket weights being trainable while the synaptic weights between input and hidden, and hidden and output neuron remain fixed.
We compute the MSE loss between the actual output and expected output per timestep. We define output per the trace $\hat{y}$ that follows from the spike output.

For BPTT with SGs, the gradient is computed as:
\begin{equation}\label{eq:delay_bptt}
\frac{\partial L}{\partial v_{j}^k}
= \frac{\partial L}{\partial \hat{y}_j(t)}
\frac{\partial \hat{y}_j(t)}{\partial s_j(t)} 
{\color{c0}{\frac{\partial s_j(t)}{\partial z_j(t)}}}
\frac{\partial z_j(t)}{\partial y_j(t)}
\frac{\partial y_j(t)}{\partial x_{j}(t)}
\left(
{\color{c1}{
\sum_{t'<t}
\frac{\partial x_{j}(t)}{\partial x_{j}(t')}
\frac{\partial x_{j}(t')}{\partial v_{j}^k}}}
+ \frac{\partial x_{j}(t)}{\partial v_{j}^k}
\right)
\end{equation}
BPTT propagates {\color{c0}back through the spikes via an SG $z_j$} and then {\color{c1}back in time through the buckets}, rather than treating the system as feedforward as in the \spikgam{} model.

\begin{figure}[!b]
    \centering
    \includegraphics[width=\linewidth]{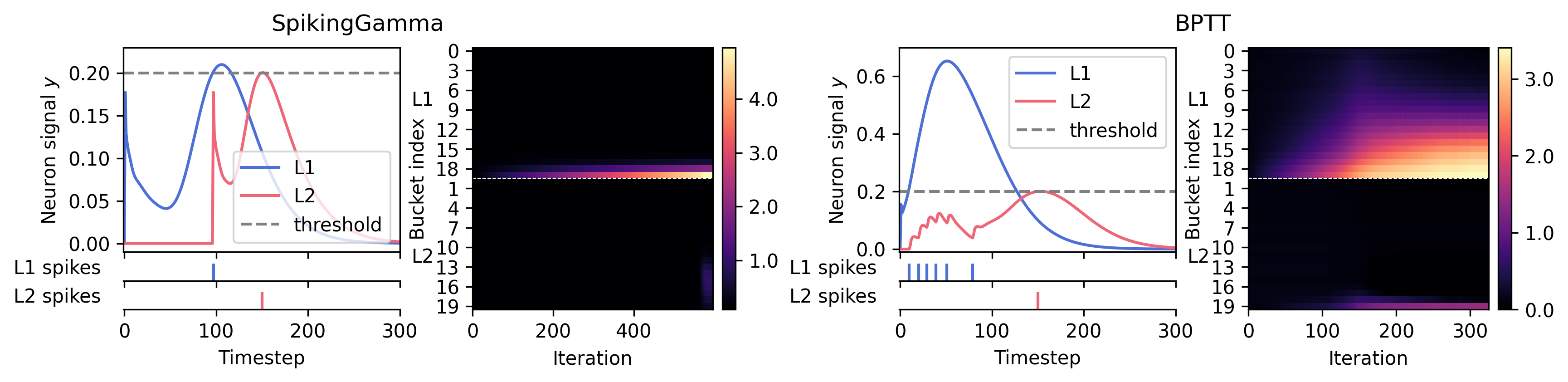}
    \caption{Comparison between \spikgam{} (left) and BPTT with SGs (right), following the backward path from (g) to (f) to (c) in Figure \ref{fig:structure_neuron_model_with_signals}. Left shows the neuron dynamics after converging, and right the bucket-weight evolution during training. Full details of the experiment setup in Appendix~\ref{tab:hyperparameters_delay_experiment}.}
    \label{fig:delay_learning}
\end{figure}

In Figure \ref{fig:delay_learning}, we show neuron signal dynamics after training (left) and bucket-weight evolution during training (right). Both approaches succeed in learning the long delay.
However, the \spikgam{} SNN can precisely target delay kernels at the timesteps where signal increases should occur, allowing it to realize the delay with minimum spikes (two), whereas BPTT with SGs reinforces a broader band of kernels, an effect that may reflect diffuse temporal credit assignment and the smooth surrogate derivative spreading gradients across nearby timesteps, consistent with \cite{li2024directly}. 

\begin{figure}[!tb]
    \centering
    \begin{minipage}{0.5\linewidth}
        \centering
        \includegraphics[width=\linewidth]{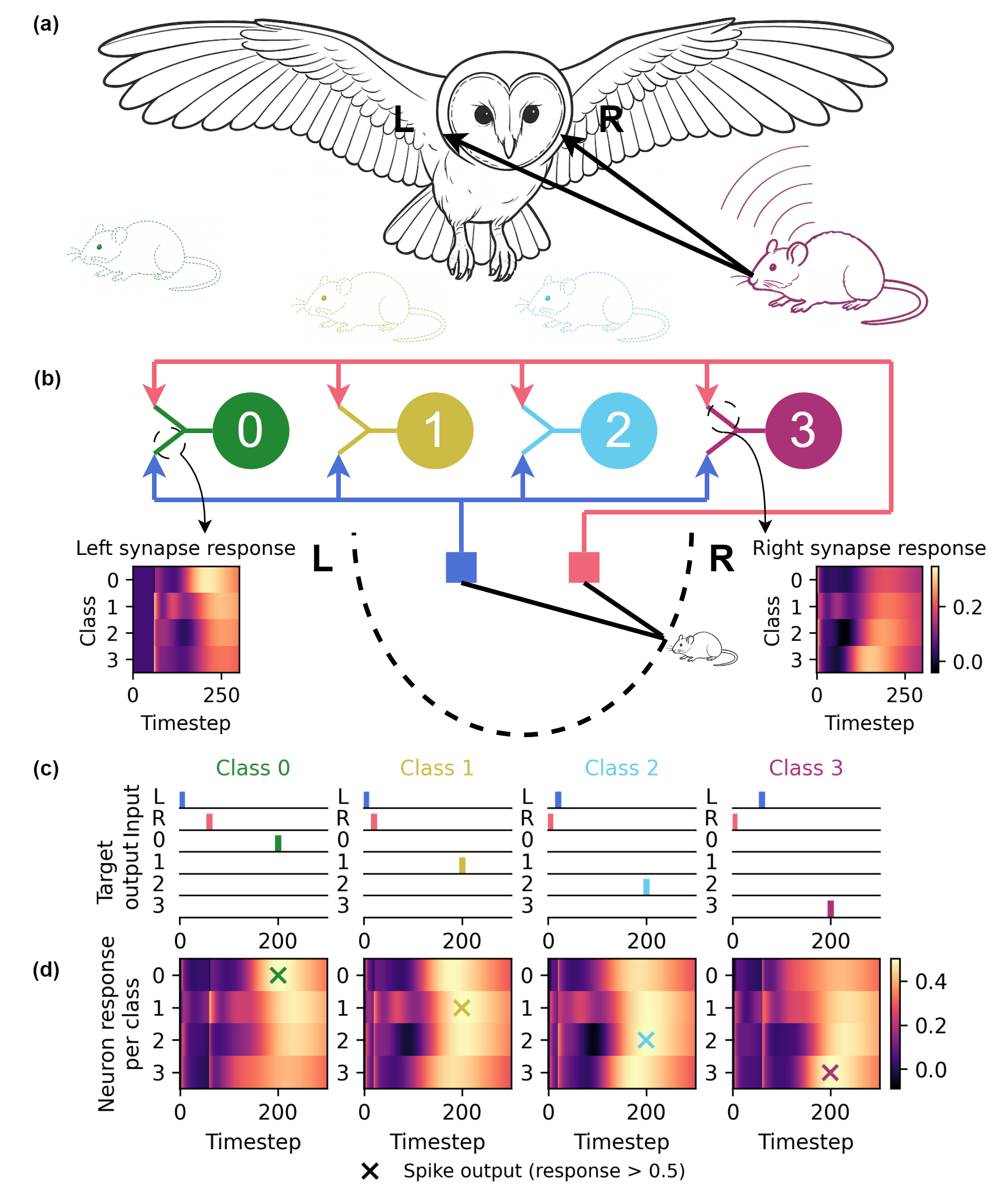}
    \end{minipage}%
    \hfill
    \begin{minipage}{0.45\linewidth}
        \caption{\textbf{(a)} An owl detects sound waves emitted by a target, such as moving prey, and these waves reach each ear at slightly different times. This difference in travel distance creates an interaural time difference, which the owl uses to accurately determine the prey’s position in the horizontal plane. \textbf{(b)} The architecture of the coincidence detection network, shown activated for a prey in the right-most (brown) position. The weighted bucket activations are shown for the left and right synapses of each class, with the left being more delayed. \textbf{(c)} Input and target spikes for all four classes. (\textbf{d}) Output neuron response to the correct class input as given in (c). Output spike (colored cross) is emitted at about $t=200$ in the correct class. Summing up the synaptic responses from (b) would result in Class 3 activation.}
        \label{fig:coincidence_detection}
    \end{minipage}
\end{figure}

\paragraph{Learning delayed coincidence detection.} In the second task, illustrated in Figure \ref{fig:coincidence_detection}, the goal is to learn delayed coincidence detection. Here, a sound arriving from a particular direction is represented by a distinct pair of spike times across two input channels, where the class is determined by the relative time-difference between the two spikes. This mimics the mechanism used by barn owls to localize prey based on subtle interaural time differences in the arrival time of spikes as a function of azimuth of the prey relative to the owl's head direction \cite{carr1990circuit}. 

The input consists of a spike pair (left (L) spike time, right (R) spike time) as defined by the class, plus jitter drawn from $\mathrm{Uniform}(0,2)$. For the four classes, the pairs are: (4, 60), (4, 20), (20, 4), and (60, 4). The target for each class is a single output spike at time $200$ from the corresponding class neuron, with other output neurons remaining silent (see Figure \ref{fig:coincidence_detection}b). The loss is computed the same way as for learning delays.
Both the left and right input have their own unique synaptic connections to all four output neurons (so 8 connections in total), thus each output neuron has two synaptic inputs (see Figure \ref{fig:coincidence_detection}a). Each of these synapses has 25 buckets, where the bucket weights are trainable and applied on synapse level. Within each neuron, the synaptic responses of both inputs are summed up.

Figure \ref{fig:coincidence_detection}c shows the synaptic responses after training. Effectively, each synapse learns a specific approximate delay such that for a specific pair of input spike times, their sum (L+R) results in a spike for the right class at the right time (see Figure  \ref{fig:coincidence_detection}d). Accuracy (per time-to-first-spike coding) is 100\% after training, with minimal spiking (maximal efficiency). This demonstration suggests that the \spikgam{} SNN is well suited for temporally structured classification tasks and aligned with biological neural coincidence detection.

\subsection{Scaling through space and time}

\begin{table*}[!t]
    \centering
    \caption{
    Comparison of online SNN training methods. Results are reported as mean $\pm$ std; primary values are final test accuracies, with peak test accuracies during training in parentheses where available. Architectures are given as [hidden neurons $\times$ hidden layers] for fully connected networks; more complex architectures are detailed in Appendix~\ref{appendix:dvsgesture_model_arch}. All results are from original papers except where noted and SSC for ES-D-RTRL, which is our reproduction. 
    Dashes denote unavailable values. Offline BPTT SG SoTa given for reference. }
    \begin{tabular}{llllll}
        \toprule
        \textbf{Task} & \textbf{Frames} & \textbf{Method} & \textbf{Architecture} & \textbf{Accuracy ± std (peak)}  \\
        \midrule
        DVS & 1000 & FPTT & 2-layer FC, LTC & - (91.28 ± 1.05) \\
        Gesture & 500 & DECOLLE & 4-layer CNN, CUBA & - (95.54 ± 0.16) \\
        & 2000 & SpikingGamma & 2-layer FC  & \textbf{92.30 ± 0.18 (93.81 ± 0.18)} \\ 
        & 2000 & SpikingGamma & 4-layer CNN & \textbf{95.08 ± 0.93 (96.08 ± 0.18)} \\ 
        \hdashline        
        \addlinespace[1.0pt]
        & {\em 20} & {\em BPTT PLIF \cite{fang2021incorporating}} & {\em 7-layer CNN} & {\em - (97.6)} \\       
        \midrule
        SHD & 50 & OTTT \cite{summe2024estimating} & $[512 \times 3]$, LIF & 71.2 ± 0.8 (-) \\
        & 50 & OSTL \cite{summe2024estimating} & $[512 \times 3]$, LIF & 70.6 ± 0.7 (-) \\
        & 50 & OTPE \cite{summe2024estimating} & $[512 \times 3]$, LIF & 75.4 ± 0.5 (-) \\
        & 100 & DECOLLE \cite{quintana2024etlp}& $[450]$, ALIF & 62.01 ± 0.61 (-) \\
        & 100 & ES-D-RTRL & $[1024 \times 3]$, RadLIF & - (93.35 ± 0.36) \\
        & 100 & e-prop \cite{quintana2024etlp} & $[450]$, ALIF & 80.79 ± 0.39 (-) \\
        & 250 & SpikingGamma & $[256 \times 3]$ & \textbf{92.81 ± 0.68 (93.55 ± 0.48)} \\
        \hdashline
        \addlinespace[1.0pt]
        & {\em 250} & {\em BPTT SE-adLIF \cite{baronig2025advancing}} & {\em $[360 \times 2]$} & {\em  93.79 ± 0.76} \\
        \midrule
        SSC & 100 & ES-D-RTRL & $[1024 \times 3]$, RadLIF & - (66.63 ± 0.53) \\
        & 250 & SpikingGamma & $[512 \times 3]$ & \textbf{75.63 ± 0.44 (75.91 ± 0.54)} \\
        \hdashline
        \addlinespace[1.0pt]
        & {\em 300} & {\em BPTT SE-adLIF \cite{baronig2025advancing}} & {\em $[720 \times 2]$} & {\em  80.44 ± 0.26} \\
        \bottomrule
    \end{tabular}
    \label{tab:sota_results}
\end{table*}

To demonstrate scaling to larger networks, problems and fine temporal resolution, we evaluate the \spikgam{} model on three widely used benchmarking datasets. We use DVS Gesture \cite{amir2017low}, SHD, and SSC \cite{cramer2020heidelberg}; while SHD and SSC can be handled with relatively shallow fully connected network topologies, they remain challenging because of the need to capture rich temporal dynamics. In contrast, DVS Gesture is less demanding temporally but has greater spatial complexity, for which we adopt both a shallow fully-connected and a deeper convolutional architecture. For SHD, we also include an ablation study to examine the effect of normalization, ReLU, and having a power-law shaped neuron response, and a study into the computational complexity and resource usage, both in Appendix~\ref{appendix:extended_results}. 
Details of the experimental setup are provided in Appendix~ \ref{appendix:exp_setup}.

As noted, online training methods seek temporal credit assignment without BPTT or RTRL. However, because existing approaches approximate temporal gradients through traces or regularizers, they may fail to capture temporal dynamics accurately. To assess their effect on accuracy and compare directly with \spikgam{} SNNs, Table \ref{tab:sota_results} reports our best results alongside prior online training methods.
We include approaches that either provide results for training with a large number of frames (DVS Gesture) or report any results at all (SHD).
We also distinguish methods reporting peak test accuracy during training, since this can introduce mild test set leakage \cite{meszaros2025efficient}. For reproducibility, we report both final-epoch and peak test accuracy.

As shown in Table \ref{tab:sota_results}, \spikgam{} SNNs achieve competitive test accuracy across datasets at superior temporal resolution. This is evident for FPTT and DECOLLE on DVS Gesture, and even more so for SHD, where most other online methods struggle. While DECOLLE performs well on DVS Gesture, it shows limited accuracy on SHD. For SHD, ES-D-RTRL is competitive provided just 100 input frames are used. Compared to offline BPTT, SpikingGamma closes much of the accuracy gap while preserving online per-timestep training and stable behavior under much finer discretization. Finally, when examining the temporal dynamics learned by \spikgam{} neurons we find they qualitatively resemble biological time-cell responses (details in Appendix~\ref{sec:time_cells}).

\paragraph{Temporal precision}
\begin{figure*}[!htb]
    \centering
    \includegraphics[width=1.0\linewidth]{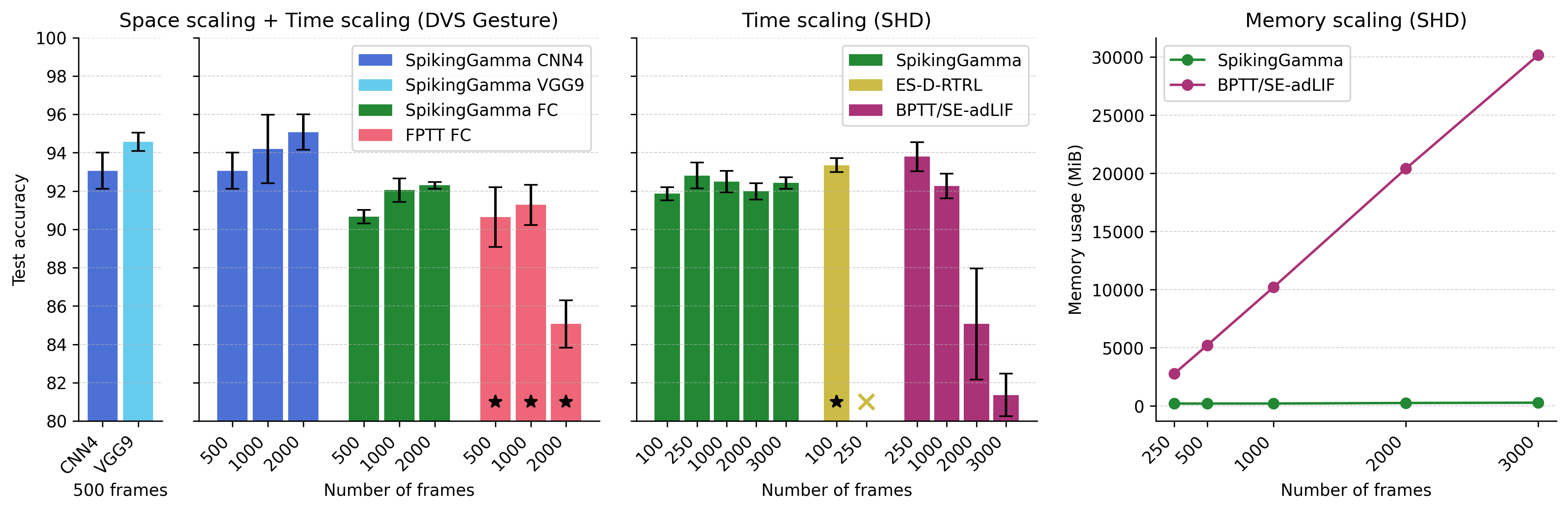}
    \caption{
    Effect of increasing the number of frames (time scaling) and model size (space scaling) on test accuracy and memory usage. The leftmost panel shows space scaling by comparing a 4-layer CNN (CNN4) with a VGG9 network at 500 frames on DVS Gesture. The remaining panels illustrate how increasing frames affects test accuracy and memory. The “$\times$” denotes chance-level performance. Bars with an asterisk (*) indicate peak test accuracy during training; others show final test accuracy after training. Detailed results, including ES-D-RTRL performance between 100 and 250 frames, are in Appendix~\ref{appendix:increase_n_frames_results}. Peak SpikingGamma results appear in Appendix~\ref{appendix:peak_training_results}. Additionally, \cite{yin2023fptt} reports that LSTM, ASRNN, and LTC neurons trained with BPTT on DVS Gesture show degrading test accuracy as frame count increases, notably between 200 and 500 frames.
    } 
    \vspace{-0.1cm}
    \label{fig:test_acc_per_dt_method}
\end{figure*}

Increasing the temporal resolution results in longer sequences, which by itself has a strongly detrimental effect on BPTT performance \cite{yin2023fptt}. Moreover, BPTT/RTRL approximations often perform well only after extensive hyperparameter tuning \cite{xue2024reasonable} and studies of achievable temporal precision are lacking except for comparatively easy datasets such as DVS Gesture \cite{wang2024brainscale, yin2023fptt}, which is known to contain limited temporal structure \cite{chen2025neuromorphic}. Here, we perform a more extensive study by evaluating performance at finer temporal resolutions (i.e., more frames of smaller time bins, with unchanged sample duration; $t$ indexes timesteps, not physical time) for both DVS Gesture and SHD. To keep the temporal dynamics consistent, we scale the bucket transfer rate by the same factor as the timestep change. To further broaden the comparison, we additionally train with a larger number of frames using publicly available implementations of FPTT, ES-D-RTRL, and SE-adLIF.

Figure \ref{fig:test_acc_per_dt_method} shows that as the number of frames increases, \spikgam{} maintains stable accuracy on DVS Gesture, even improving slightly, while FPTT degrades sharply. On SHD, ES-D-RTRL collapses, and SE-adLIF also deteriorates substantially, whereas \spikgam{} maintains performance. Memory usage remains low and theoretically constant over time, increasing slightly in practice due to software overhead, while BPTT-based methods scale linearly with the number of frames. The leftmost panel further shows that \spikgam{} scales effectively to larger and deeper models (from CNN4 to VGG9), resulting in improved accuracy.

\vspace{-0.1cm}
\paragraph{Sparsity}
\spikgam{} SNNs can theoretically achieve efficient spike-coding, as shown by the coincidence detection task, by keeping dynamics mostly subthreshold and using spike-time codes rather than relying solely on rate codes. This is similar to how delays or adaptive neurons enable efficient neural codes. When applying gain loss on SHD, we reach a spike density comparable to competitive methods (5 to 6 spikes/neuron/sample \cite{deckers2024co} with little accuracy loss), as shown in Figure \ref{fig:spike_density} (Appendix). However, because SHD is essentially rate-coded \cite{cramer2020heidelberg}, its potential for sparsity is inherently limited \cite{yu2025beyond}. For datasets with richer temporal structures, we expect larger gains in sparsity.

\section{Discussion and Conclusion}

We introduce the \spikgam{} model, an online training approach for SNNs that enables direct error backpropagation through the sigma-delta step using a straight-through estimator, without needing surrogate gradients or temporal unrolling. The method thereby avoids the temporal credit assignment approximations inherent to existing online methods.
This allows \spikgam{} to learn at fine temporal resolutions while preserving the capacity to capture complex temporal dependencies across multiple timescales. By maintaining information through subthreshold dynamics, the model also encourages sparse spike coding, improving compatibility with the communication constraints of neuromorphic hardware.

Further sparsity may be achievable by exploiting \spikgam’s ability to learn complex temporal feature detectors at both the neuron and synapse level. However, many large-scale benchmarks are defined by rate-code models, making it difficult to evaluate and benefit from sparse temporal coding strategies \cite{yu2025beyond}. Benchmarks designed around temporal coding could therefore better reveal the efficiency gains enabled by \spikgam.

At the same time, \spikgam{} introduces additional state and computation through its bucket representation, with costs scaling with the number of buckets $K$. Moreover, while \spikgam{} avoids BPTT in feedforward architectures, BPTT-like training remains necessary for networks with recurrent connections or other feedback pathways. Finally, although our software results support temporal precision and stable online training, the expected efficiency benefits depend on hardware support for local bucket dynamics and still require validation on neuromorphic substrates. With modern AI dominated by feedforward architectures such as Transformers, analogous extensions of \spikgam{} may offer a path toward scalable deep SNNs for sequence learning.

\section*{Acknowledgments} 
Roel Koopman is funded by the Dutch Research Council under Grant agreement KICH1.ST04.22.021. Sander Bohté is supported by NWO-NWA grant NWA.1292.19.298. Sebastian Otte acknowledges support from the Alexander von Humboldt Foundation (Feodor Lynen Research Fellowship) during the early phase of this work.

\bibliographystyle{unsrtnat}
\bibliography{references}

\newpage
\appendix
\onecolumn
\counterwithin{figure}{section}
\counterwithin{table}{section}
\counterwithin{equation}{section}
\renewcommand{\thefigure}{\Alph{section}\arabic{figure}}
\renewcommand{\thetable}{\Alph{section}\arabic{table}}
\renewcommand{\theequation}{\Alph{section}\arabic{equation}}

\section{Pseudocode}\label{appendix:pseudocode}
\begin{algorithm}[H]
\caption{SpikingGamma neuron forward pass for a single timestep from $t-1$ to $t$}
\label{alg:pseudocode}
\small
\begin{algorithmic}
\Require current bucket state of previous layer $\mathbf{\hat{y}}_{\mathrm{in}}[t] \in \mathbb{R}^{N_{\mathrm{in}} \times K}$, previous bucket state of current layer $\mathbf{\hat{y}}_{\mathrm{out}}[t-1] \in \mathbb{R}^{N_{\mathrm{out}} \times K}$, synaptic weights $\mathbf{W} \in \mathbb{R}^{N_{\mathrm{in}} \times N_{\mathrm{out}}}$, bucket weights $\mathbf{V} \in \mathbb{R}^{N_{\mathrm{out}} \times K}$, spike threshold $\vartheta[t] \in \mathbb{R}_{>0}$, shared bucket transfer rates $\boldsymbol{\alpha} \in \mathbb{R}^{K}$
\Ensure updated bucket state $\mathbf{\hat{y}}_{\mathrm{out}}[t] \in \mathbb{R}^{N_{\mathrm{out}} \times K}$, output spikes $\mathbf{s}[t] \in \{0,1\}^{N_{\mathrm{out}}}$
\State $\mathbf{x}[t] \gets \sum_{k} \left((\mathbf{\hat{y}_\text{in}}[t]^T \mathbf{W})^T \circ \mathbf{V}\right)_{:,k}$
\State $\mathbf{y}[t] \gets \mathrm{ReLU}(\mathbf{x}[t])$
\State $\mathbf{z}[t] \gets \mathbf{y}[t] - \sum_k\mathbf{\hat{y}}_{\mathrm{out}}[t-1]_{:,k}$
\State $\mathbf{s}[t] \gets \mathbf{z}[t] > \vartheta[t]$

\State $\mathbf{\hat{y}}_{\mathrm{out}}[t]_{:,0} \gets \alpha_0 \mathbf{\hat{y}}_{\mathrm{out}}[t-1]_{:,0} + 2\vartheta[t]\mathbf{s}[t]$

\For{$k=1$ to $K-1$}
    \State $\mathbf{\hat{y}}_{\mathrm{out}}[t]_{:,k} \gets \alpha_k \mathbf{\hat{y}}_{\mathrm{out}}[t-1]_{:,k} + (1-\alpha_k)\mathbf{\hat{y}}_{\mathrm{out}}[t-1]_{:,k-1}$
\EndFor
\end{algorithmic}
\end{algorithm}

\section{SpikingGamma SSM interpretation}
\label{math:ssm}
SpikingGamma can be interpreted within the State-Space Model (SSM) framework. 
For layer $\ell$, we stack all bucket states into 
$\mathbf{h}^{\ell}(t)=\big[\hat{y}_i^{k,\ell}(t)\big]_{i,k}$, yielding
\begin{equation}
\mathbf{h}^{\ell}(t) = \mathbf{A}\,\mathbf{h}^{\ell}(t-1) + \mathbf{B}\,\mathbf{s}^{\ell}(t), \qquad \mathbf{D}=0
\end{equation}
where $\mathbf{A}$ implements the Gamma bucket dynamics and $\mathbf{B}$ injects spike events into the first bucket (cf.~Eq.~\eqref{eq:bucket_dynamics}, ignoring the adaptive threshold). 
The model has two readouts,
\begin{equation}
\hat{\mathbf{y}}^{\ell}(t) = \mathbf{C}_{\mathrm{out}}\,\mathbf{h}^{\ell}(t), \qquad
\mathbf{y}^{\ell}(t) = \phi\!\left(\mathbf{C}_{\mathrm{in}}\,\mathbf{h}^{\ell-1}(t)\right)
\end{equation}
where $\mathbf{C}_{\mathrm{out}}$ just sums buckets per neuron, while 
$\mathbf{C}_{\mathrm{in}}$ also applies $w_{ij}$ and $v_j^k$, and 
$\phi(\cdot)=\mathrm{ReLU}(\cdot)$.

The state is driven by $\mathbf{s}^{\ell}(t)$, a quantized representation of the continuous drive $\mathbf{y}^{\ell}(t)$. 
Spikes are generated by elementwise thresholding the prediction error:
\begin{equation}
\mathbf{s}^{\ell}(t)
= (\mathbf{y}^{\ell}(t) - \hat{\mathbf{y}}^{\ell}(t-1) > \vartheta(t)).
\end{equation}
Each spike thus both corrects the local reconstruction via $\hat{\mathbf{y}}^\ell$ and drives the next layer via $\mathbf{y}^{\ell + 1}$.

\section{Extended results}\label{appendix:extended_results}
\subsection{Performance comparison when increasing number of frames}\label{appendix:increase_n_frames_results}
For the reproduced works, we examined how time constants were applied and whether they required adjustment when the number of frames (and thus the timestep size) changed.
For FPTT, ES-D-RTRL, and SE-adLIF, the time constants are trainable. We observed no performance improvements from modifying their bounds or initialization.

\begin{table}[H]
    \centering
    \caption{Test accuracy (peak for FPTT, final for SpikingGamma) for models trained on DVS Gesture. *Reproduced from \url{https://github.com/byin-cwi/sFPTT}. No annotation means the number is from the original work.}
    \begin{tabular}{lllll}
    \toprule
    \textbf{Frames} & \textbf{SpikingGamma} & \textbf{SpikingGamma} & \textbf{SpikingGamma} & \textbf{FPTT} \\
    & \textbf{(CNN4)} & \textbf{(VGG9} & \textbf{(FC)} & \\
    \midrule
    500 & 93.06 ± 0.94 & 94.57 ± 0.47 & 90.66 ± 0.36 & 90.64 ± 1.56 \\
    1000 & 94.19 ± 1.79 & - & 92.05 ± 0.62 & 91.28 ± 1.05 \\
    2000 & 95.08 ± 0.93 & - & 92.30 ± 0.18 & 85.07 ± 1.24* \\
    \bottomrule
    \end{tabular}
    \label{tab:exres_dvsgesture_fc}
\end{table}

\begin{table}[H]
    \centering
    \caption{Test accuracy (peak for ES-D-RTRL, final for SpikingGamma and BPTT/SE-adLIF) for models trained on SHD. *Reproduced from \url{https://github.com/chaobrain/brainscale-exp-for-snns} (commit 11720f5). The number of epochs was reduced from 100 to 30. For 250+ frames, ES-D-RTRL no longer converges, see Figure \ref{fig:esdrtrl_100to300frames} for transition and collapse from 100 to 300 frames. Therefore, we did not reproduce for 1000 or more frames. **Reproduced from \url{https://github.com/IGITUGraz/SE-adlif}. No annotation means the number is from the original work.}
    \begin{tabular}{llll}
    \toprule
    \textbf{Frames} & \textbf{SpikingGamma} & \textbf{ES-D-RTRL} & \textbf{BPTT/SE-adLIF} \\
    \midrule
    100 & 91.86 ± 0.34 & 93.35 ± 0.36 & - \\
    250 & 92.81 ± 0.68 & 4.5* & 93.79 ± 0.76 \\
    1000 & 92.49 ± 0.57 & - & 92.26 ± 0.65** \\
    2000 & 91.99 ± 0.43 & - & 85.06 ± 2.89** \\
    3000 & 92.42 ± 0.31 & - & 81.36 ± 1.11** \\
    \bottomrule
    \end{tabular}
    \label{tab:exres_shd}
\end{table}
\begin{figure}[H]
    \centering
    \includegraphics[width=0.4\linewidth]{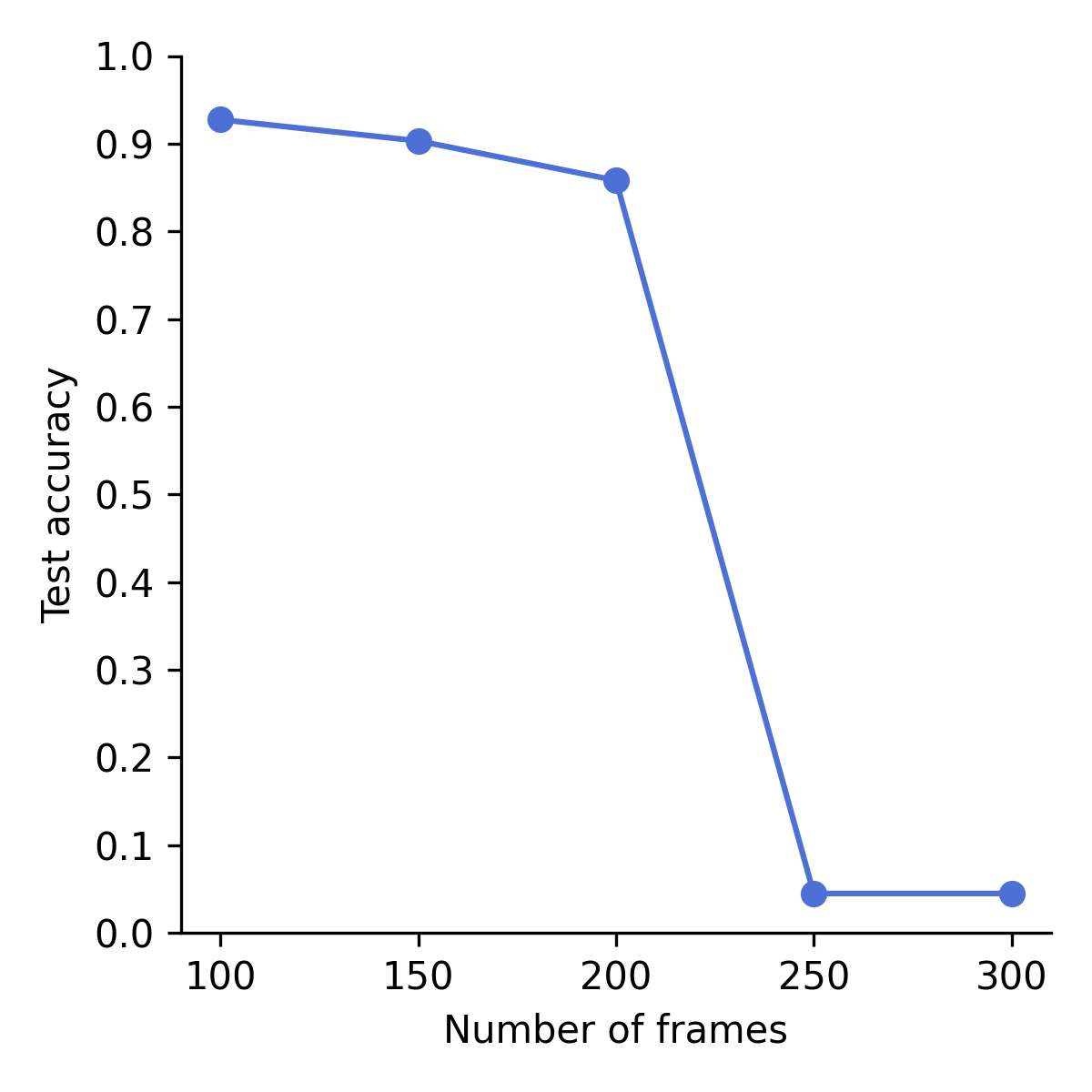}
    \caption{Test accuracy (best over all epochs) for ES-D-RTRL (reproduced) trained on SHD from 100 to 300 frames. The accuracy for 250 frames was verified over 3 independent training sessions.}
    \label{fig:esdrtrl_100to300frames}
\end{figure}

\subsection{Peak test accuracy during training}\label{appendix:peak_training_results}
A common practice in prior work is to select the peak test performance observed during training, effectively using the test set to guide model selection. This introduces test set information into the training process and can lead to fitting on the test set, thereby overestimating generalization performance. To avoid this issue, we did not adopt this practice. However, to ensure a fair comparison with related works that did use this strategy, in Table \ref{tab:peak_test_during_training} we additionally report the peak test performance achieved over the course of training (with the number of epochs increased to 30 for SHD/SSC and to 150 for DVS Gesture). As expected, these numbers are slightly higher, but the overall trends remain consistent.

\begin{table}[H]
    \centering
    \caption{Peak test accuracies achieved during training.}
    \begin{tabular}{lll}
    \toprule
    \textbf{Task} & \textbf{Frames} & \textbf{Accuracy ± std} \\
    \midrule
    DVS Gesture (CNN4) & 500 & 95.58 ± 0.18 \\
     & 1000 & 96.21 ± 0.31 \\
     & 2000 & 96.08 ± 0.18 \\
     \midrule
     DVS Gesture (VGG9) & 500 & 95.71 ± 0.36 \\
     \midrule
    DVS Gesture (FC) & 500 & 92.80 ± 0.62 \\
     & 1000 & 93.31 ± 0.18 \\
     & 2000 & 93.81 ± 0.18 \\
     \midrule
    SHD & 100 & 92.64 ± 0.27 \\
     & 250 & 93.55 ± 0.48 \\
     & 1000 & 93.18 ± 0.34 \\
     & 2000 & 92.61 ± 0.39 \\
     & 3000 & 93.31 ± 0.24 \\
     \midrule
    SSC & 250 & 75.91 ± 0.54 \\
    \bottomrule
    \end{tabular}
    \label{tab:peak_test_during_training}
\end{table}

\subsubsection{Clarifying reported test accuracy for DECOLLE}\label{appendix:reproduce_decolle}
Since the original work on DECOLLE in \cite{kaiser2020synaptic} was ambiguous about whether it reported peak or final epoch test accuracy, we reproduced their results using the original code (\url{https://github.com/nmi-lab/decolle-public}) to clarify this point. Across four independent runs, we obtained a peak test accuracy of 95.70 ± 0.47 and a final epoch test accuracy of 94.19 ± 0.94. The peak accuracy aligns much more closely with the reported value of 95.54 ± 0.16, indicating that the original paper most likely reported peak test accuracy rather than final epoch accuracy.

\subsection{Effect of introducing a variable threshold}\label{appendix:variable_threshold_result}
\begin{figure}[H]
    \centering
    \includegraphics[width=0.4\linewidth]{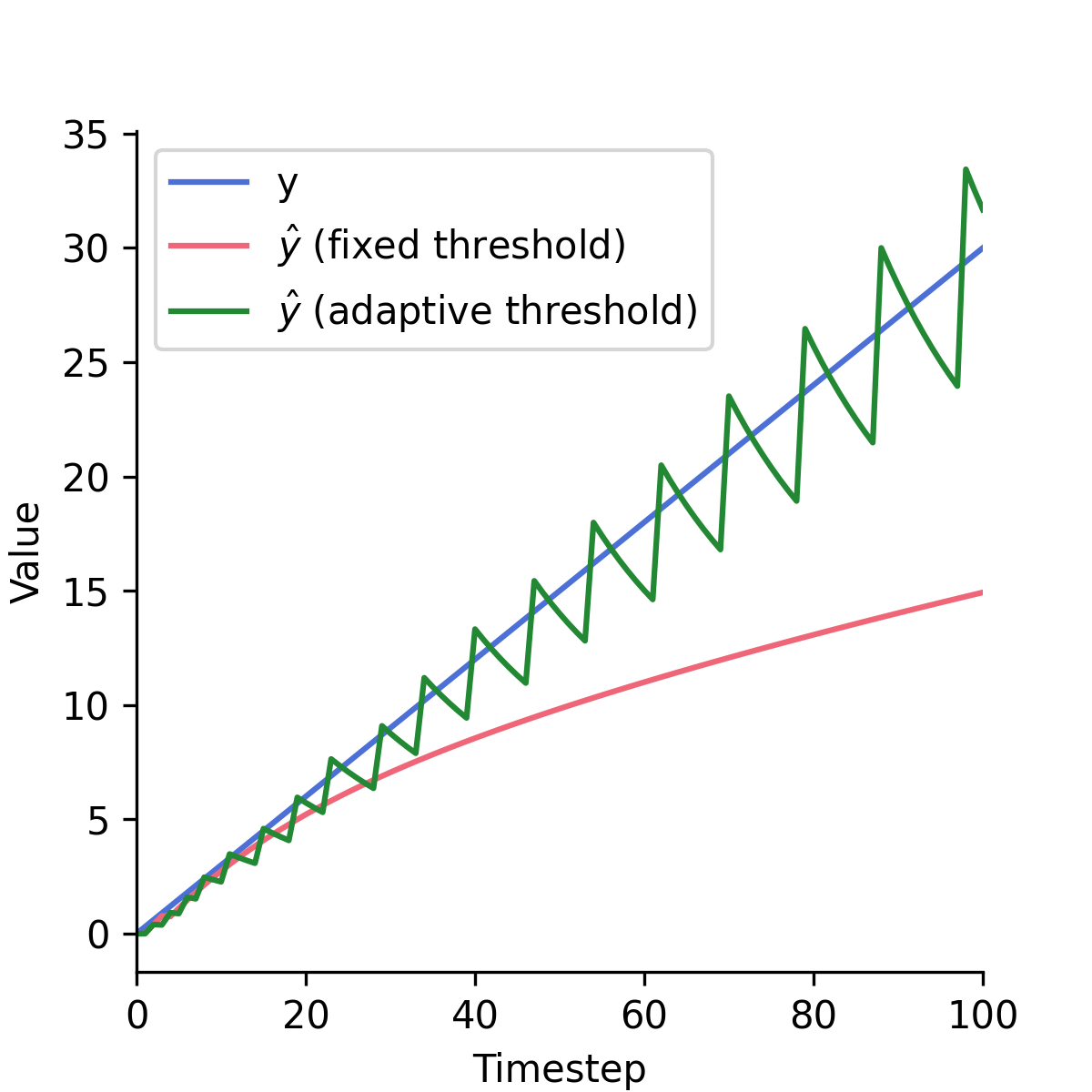}
    \caption{A demonstration showing that, without adaptive thresholding, $\hat{y}$ is unable to accurately estimate higher values of $y$, whereas the use of adaptive thresholding enables $\hat{y}$ to effectively estimate these higher values.}
    \label{fig:adaptive_thresholding_range_increase}
\end{figure}

\subsection{ES-D-RTRL result for SSC}\label{appendix:esdrtrl_ssc}
The SSC results are based on reproduced work from \url{https://github.com/chaobrain/brainscale-exp-for-snns} (commit 11720f5). We ran 4 training sessions with the same hyperparameters as they used for their best SHD results, and then reported the mean and standard deviation of the peak epoch of each run.

\subsection{Alignment with biological time-cell dynamics}\label{sec:time_cells}
\begin{figure}[H]
    \centering
    \includegraphics[width=0.7\linewidth]{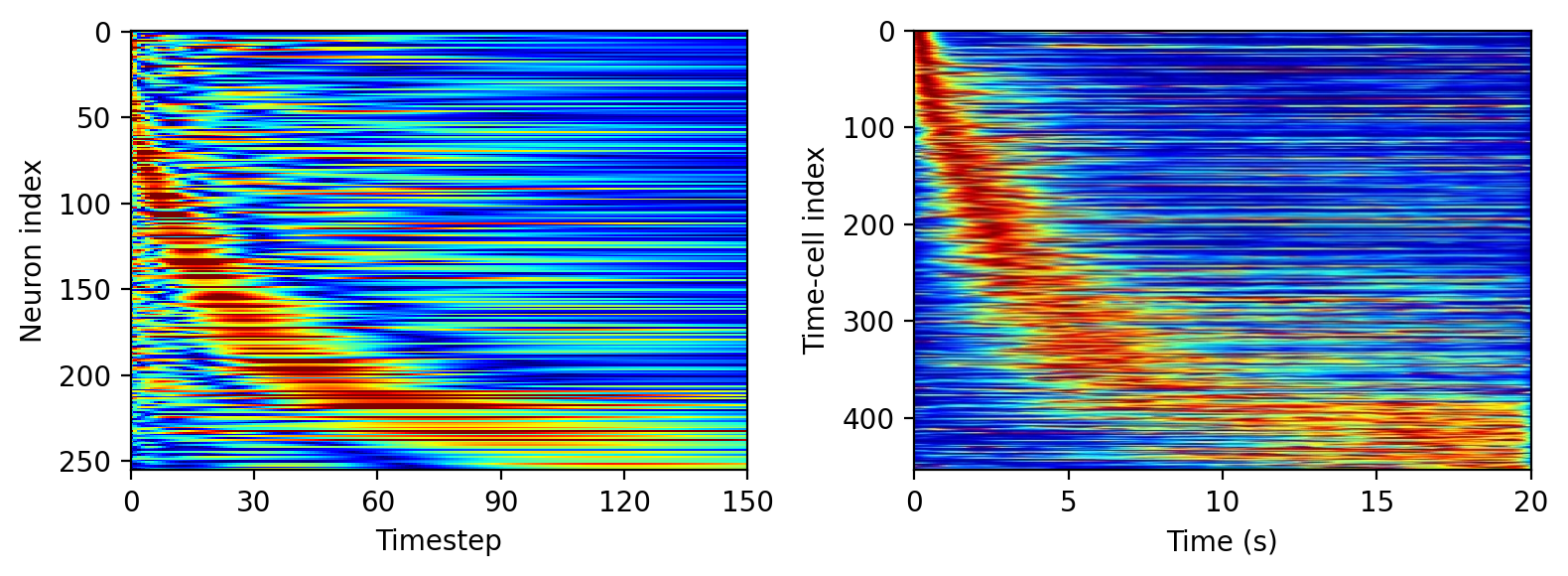}
    \caption{Comparison of neuron responses from our method with recorded rat time-cell activity. On the left: Neuron response for input at $t=0$ for neurons from the second hidden layer of the best performing model on SHD. Neuron indices sorted by peak activity timestep. On the right: Firing patterns of rats’ time-cells during a temporal bisection task. Adapted from \cite{shimbo2021scalable}.}
    \label{fig:compare_with_time_cells}
\end{figure}

To examine whether neurons in our model show time-cell-like dynamics, we analyzed the responses of neurons in a single layer to a single spike input after training on SHD. Comparing these neuronal responses to recorded biological time-cell activity revealed similar population dynamics. In both cases, the units activate sequentially over time with gradually decaying and temporally broadened responses, forming a characteristic power-law-like pattern (Figure \ref{fig:compare_with_time_cells}). This suggests that our model captures temporal information in a manner consistent with biological neural representations of time.

\subsection{Ablation study on SHD}\label{sec:ablation_study_shd}
To gather more insight into the effects of normalization, activation, and hyperparameters, we performed an ablation study on SHD. We used the same experimental setup as for our best-performing model, but introduced the modifications listed in Tables \ref{tab:ablation_results} and \ref{tab:results_alternative_kernels}.

During our experiments, the neuron signal was first normalized using layer normalization and then rectified. As shown in Table \ref{tab:ablation_results}, removing the ReLU activation used to compute the neuron signal $y$ leads to a clear drop in performance. This is expected, as negative neuron signal values cannot be represented by spikes in our method, resulting in a mismatch between the signal $y$ and its estimate $\hat{y}$.

Additionally, altering or removing this normalization can have substantial impact on accuracy. RMS normalization \cite{zhang2019root} performs similarly to layer normalization, which is expected since both normalize using the same underlying statistics. In contrast, batch normalization \cite{ioffe2015batch} completely fails. This aligns with previous findings that batch normalization is difficult to apply to the hidden-to-hidden transitions of RNNs \cite{laurent2016batch}. Batch normalization through time \cite{kim2021revisiting} was designed to better capture temporal characteristics than standard batch normalization, and our results confirm this, though its performance still lags behind layer and RMS normalization. Finally, omitting normalization altogether yields mixed results: when bucket weights are assigned to neurons, performance is poor, but assigning them to synapses improves results, though still with much higher variance.
In summary, properly normalizing neuron signal dynamics is essential for stable training and good performance when scaling to a more complex task like SHD.

\begin{table}[H]
    \centering
    \caption{Ablation study results for SHD.}
    \begin{tabular}{lll}
    \toprule
    \textbf{Ablated} & \textbf{Accuracy ± std} \\
    \midrule
    No normalization, bucket weights on neuron & 50.83 ± 9.10 \\
    No normalization, bucket weights on synapse & 88.31 ± 3.15 \\
    RMS normalization & 92.54 ± 0.63 \\
    Batch normalization & 16.59 ± 3.28 \\
    Batch normalization through time & 90.57 ± 0.60 \\
    No ReLU & 85.38 ± 1.37 \\
    \bottomrule
    \end{tabular}
    \label{tab:ablation_results}
\end{table}

In Section \ref{sec:param_bucket_transfer_rate}, we introduced the standard method for initializing the bucket transfer rates. This approach is based on the idea that a well-chosen power-law-like shape (given by the sum of all buckets) is most effective at retaining information. We test this hypothesis by using different shapes and rates, as shown in Table \ref{tab:results_alternative_kernels}. The results confirm that a power-law sum of responses, decaying over an appropriate time interval, yields the best performance, while all other response shapes underperform.

\begin{table}[H]
    \centering
    \caption{Results from experiments with alternative kernel parameters and shapes. The "Power-law" shape type refers to the default initialization method as introduced in section \ref{sec:param_bucket_transfer_rate}, while "Fixed rate" refers to an alternative scheme where $\alpha_k$ is fixed to one value (given as "Rate") for all $K$ kernels. The visualized neuron responses show the sum of all kernel responses, with on the x-axis 100 timesteps and on y-axis values from 0 to 1.}
    \begin{tabular}{lllll}
    \toprule
    \textbf{Shape type} & \textbf{Rate} / $\boldsymbol{F}$ & $\boldsymbol{K}$ & \textbf{Neuron response} & \textbf{Accuracy} \\
    \midrule
    Fixed rate & 0.82 & 10 & \includegraphics[width=0.1\linewidth]{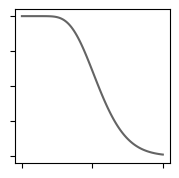} & 75.45 ± 1.16 \\
    Fixed rate & 0.15 & 10 & \includegraphics[width=0.1\linewidth]{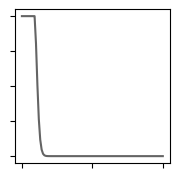} & 67.04 ± 1.48 \\
    Power-law & 1.00 & 1 & \includegraphics[width=0.1\linewidth]{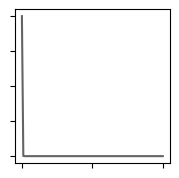} & 32.82 ± 1.09 \\
    Power-law & 0.00 & 1 & \includegraphics[width=0.1\linewidth]{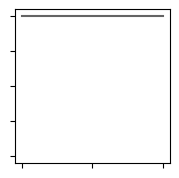} & 22.14 ± 4.71 \\
    Power-law & 0.90 & 10 & \includegraphics[width=0.1\linewidth]{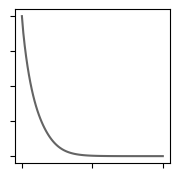} & 84.49 ± 0.88 \\
    \textbf{Power-law} & \textbf{0.15} & \textbf{10} & \includegraphics[width=0.1\linewidth]{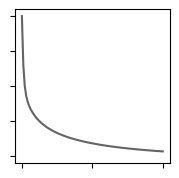} & \textbf{92.81 ± 0.68} \\
    Power-law & 0.01 & 10 & \includegraphics[width=0.1\linewidth]{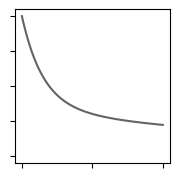} & 87.52 ± 0.38 \\
    \bottomrule
    \end{tabular}
    \label{tab:results_alternative_kernels}
\end{table}

\begin{figure}[H]
    \centering
    \includegraphics[width=0.7\linewidth]{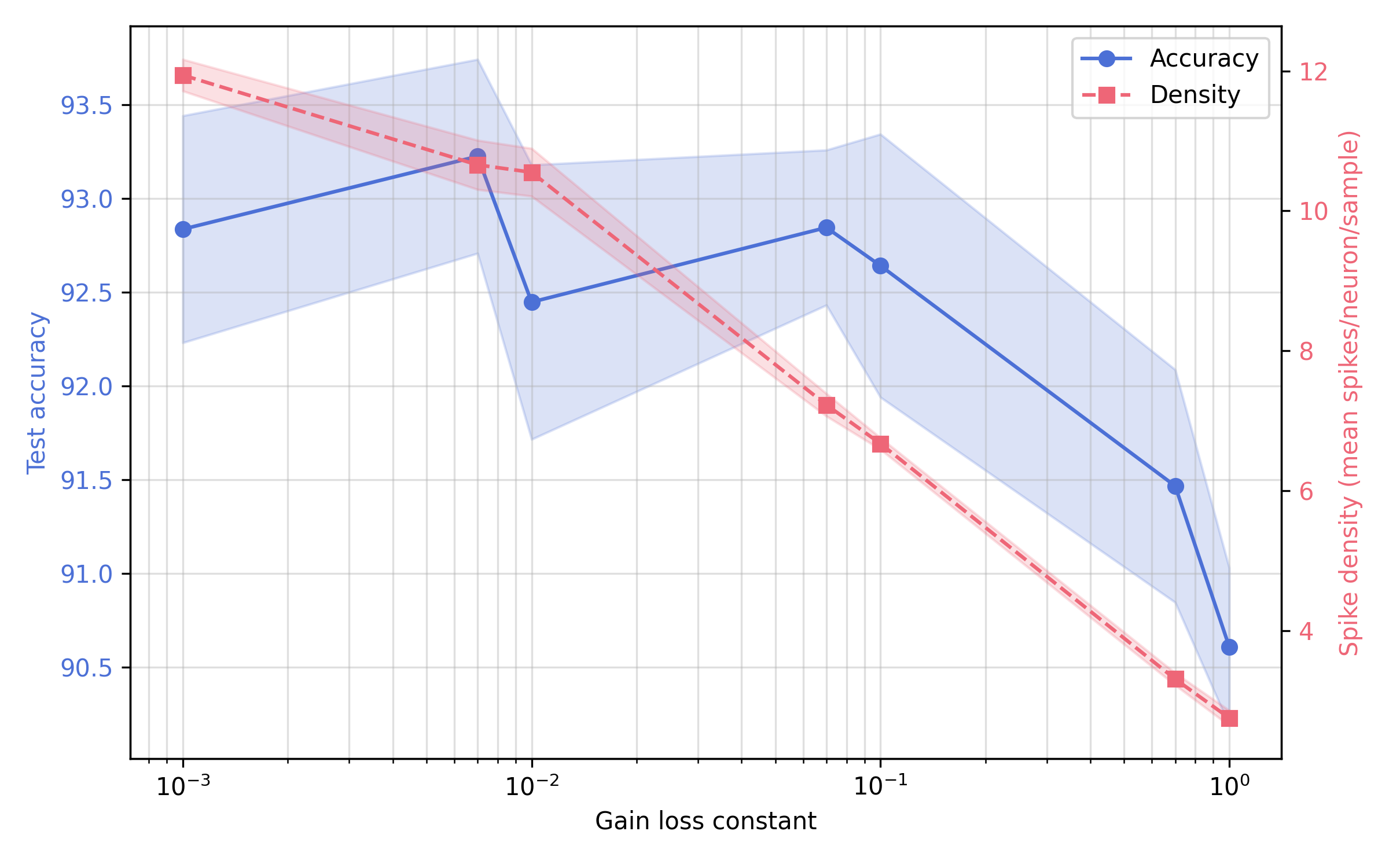}
    \caption{Effect of the gain-loss constant on accuracy and spike density (shown as mean ± std). The spike density is computed using Eq. \eqref{eq:spike_density}.}
    \label{fig:spike_density}
\end{figure}

We also experimented with the gain–loss constant to achieve sparsity, as shown in Figure \ref{fig:spike_density}. Setting the constant to 0.1 resulted in a significant increase in sparsity with little to no degradation in accuracy. Increasing the constant further continued to improve sparsity, but at the cost of progressively larger accuracy losses.

\subsection{Computational complexity}\label{appendix:computational_complexity}
The computational cost of SNNs can be split into spike- and timestep-dependent operations. Neuronal updates occur at every timestep and require MACs, while synaptic operations are spike-driven and typically consist of cheaper ACs.

In the following table, we compare with the $\Sigma\Delta$ neuron \cite{nair2019ultra} and the LIF neuron. The $\Sigma\Delta$ and LIF neuron require MACs for leakage and ACs for current accumulation. SpikingGamma follows a similar structure, but also scales with the number of buckets $K$. The threshold is fixed.

\begin{table}[H]
\centering
\caption{Computational scaling \textbf{per layer}. A layer has $N$ neurons, each with $K$ buckets, and modeled for $T$ timesteps during which there were $S$ spikes (in/out).}
\small
\begin{tabular}{llll}
\toprule
\textbf{Where} & \textbf{Operation} & \textbf{Scaling (ACs)} & \textbf{Scaling (MACs)} \\
\midrule

Synapse 
& Current addition to first input bucket 
& $N S_\text{in}$ 
& -- \\

Neuron (Gamma) 
& Leak over input buckets 
& -- 
& $T N K$ \\

& Apply bucket weights 
& -- 
& $T N K$ \\

& Sum weighted input buckets 
& $T N (K-1)$ 
& -- \\

Neuron ($\Sigma\Delta$) 
& Sum $\Sigma\Delta$ buckets 
& $T N (K-1)$ 
& -- \\

& Compute difference signal and estimate 
& $T N$ 
& -- \\

& Add spikes to first bucket 
& $S_\text{out}$ 
& -- \\

& Leak over $\Sigma\Delta$ buckets 
& -- 
& $T N K$ \\

\midrule
\textbf{Total (SpikingGamma)} 
& 
& \makecell[l]{$O(N(TK + S_\text{in})$ \\ $+\, S_\text{out})$} 
& $O(T N K)$ \\

\midrule

\textbf{Total ($\Sigma\Delta$ neuron \cite{nair2019ultra})} 
& 
& \makecell[l]{$O(N(T + S_\text{in})$ \\ $+\, S_\text{out})$} 
& $O(T N)$ \\

\textbf{Total (LIF neuron)} 
& 
& $O(N S_\text{in})$ 
& $O(T N)$ \\

\bottomrule
\end{tabular}
\label{tab:scaling}
\end{table}

With many timesteps, neuron updates dominate, making MACs more significant. Below we report ACs and MACs per SHD sample.

\begin{table}[H]
\centering
\caption{Estimated computational cost per sample for SHD. The network consists of 3 layers of each 256 neurons plus an output layer of 20 neurons, with each neuron having 10 buckets. Each sample has 250 timesteps. } 
\small
\begin{tabular}{llll}
\toprule
\textbf{Where} & \textbf{Operation} & \textbf{ACs} & \textbf{MACs} \\
\midrule

Synapse 
& Current addition to first input bucket 
& 1490K $\pm$ 165.7K 
& 0 \\

Neuron (Gamma) 
& Leak over input buckets 
& 0 
& 1970K \\

& Apply bucket weights 
& 0 
& 1970K \\

& Sum weighted input buckets 
& 1773K 
& 0 \\

Neuron ($\Sigma\Delta$) 
& Sum $\Sigma\Delta$ buckets 
& 1773K 
& 0 \\

& Compute difference signal and estimate 
& 197K 
& 0 \\

& Add spikes to first bucket 
& 5K $\pm$ 0.5K 
& 0 \\

& Leak over $\Sigma\Delta$ buckets 
& 0 
& 1970K \\

\midrule
\textbf{Total (SpikingGamma)} 
& 
& 5239K $\pm$ 166.2K 
& 5910K \\

\bottomrule
\end{tabular}
\label{tab:compute}
\end{table}

We can approximately extrapolate these numbers to the $\Sigma\Delta$ neuron by dividing terms scaling with $K$ by 10, resulting in relatively more ACs than MACs. Under standard digital assumptions, this suggests that the additional cost of SpikingGamma is dominated by timestep-dependent bucket updates. However, this should be interpreted as a rough indication, as actual energy consumption depends on hardware, precision, and memory access patterns. In mixed-signal systems, bucket operations may be significantly cheaper than MAC counts suggest.

\subsection{Sigma-delta approximation quality}\label{appendix:estimator_analysis}
\begin{figure}[H]
    \centering
    \includegraphics[width=0.5\linewidth]{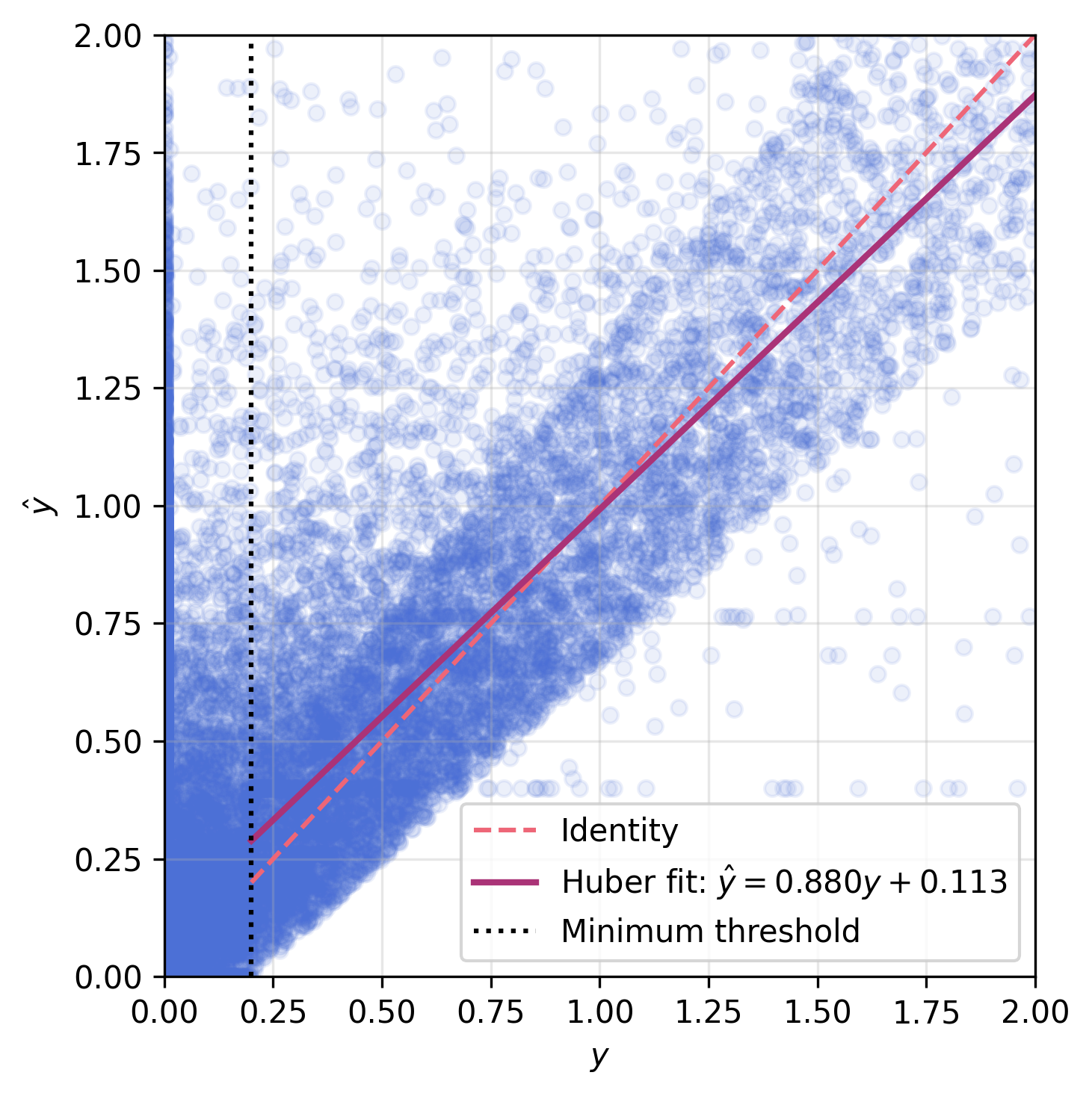}
    \caption{Empirical relationship between the activation $y$ and its approximation $\hat{y}$ for the neurons of all three hidden layers on the SHD test set after training. Each point is one neuron at one time step of one test sample (10000 random points per layer). The Huber fit uses only points with $y$ at or above the minimum spike threshold (dotted line). Hyperparameters as in Table~\ref{tab:hyperparameters}.}
    \label{fig:sd_approximation_quality}
\end{figure}

In Figure~\ref{fig:sd_approximation_quality}, we visualize the empirical relation between $y$ and $\hat{y}$. Most points lie close to the identity line, showing a good overall match. A robust fit on the points with $y \geq \vartheta_0$ has a slope slightly below one. Because the threshold grows with $\hat{y}$, both the error tolerated before a spike and the size of each spike grow with $\hat{y}$, which widens the spread for larger values. Points far above the identity line occur when $y$ drops faster than $\hat{y}$ can follow: only positive spikes are emitted, so $\hat{y}$ can only decrease through the decay of the kernel. Points far below the identity line occur when $y$ rises quickly: at most one spike is emitted per time step, so $\hat{y}$ needs several time steps to catch up. When $y$ is below the minimum threshold $\vartheta_0$, no spike can be emitted, so $\hat{y}$ only contains the decaying residual of earlier spikes, or is zero.

\section{Details on the experimental setup}\label{appendix:exp_setup}
\subsection{Datasets}\label{appendix:datasets}
The Spiking Heidelberg Digits (SHD) and Spiking Speech Commands (SSC) datasets both consist of auditory recordings. Each recording was converted into spike trains using a biologically inspired cochlear model, capturing fine-grained temporal structure in the audio \cite{cramer2020heidelberg}. 
The SHD dataset contains 8156 training and 2264 test samples for 20 spoken digits.
The SSC dataset contains 75466 training, 9981 validation, and 20382 test samples for 35 spoken commands. We included the validation samples in the training set (and did no validation during training).
For both datasets, the 700-channel cochlear outputs are downsampled to 140 channels to reduce input dimensionality.

The DVS Gesture dataset \cite{amir2017low} contains event-based visual recordings of 11 hand and arm gestures captured using a DVS camera. It includes 1176 training and 288 test samples. The original $128 \times 128$ event frames are downsampled to $32 \times 32$ by summing up events in a $4 \times 4$ window.

To prepare the event streams from the datasets for model input, the timestamps are discretized into time-bins along the temporal axis. All events falling within the same time-bin are accumulated at their corresponding spatial indices, producing a sequence of frames. These frames are then fed to the model as inputs at successive discrete timesteps.

For computing the mean and standard deviation of the reported accuracies, we repeated the experiments with SHD and SSC 5 times, and with DVS Gesture 3 times.

\subsection{Network architectures and hyperparameters}\label{appendix:datasets_network_archs}

\begin{table}[!htb]
    \centering
    \caption{Network architecture and hyperparameters. The architecture is given as [neurons in hidden layers $\times $ number of hidden layers] if fully-connected. Whenever we change the number of frames / the timestep size from the values mentioned above, we scale the bucket transfer rate by the same factor as the timestep size change to keep the temporal dynamics consistent. *Times $K$ per layer upward starting from the output towards the input.}
    \begin{tabular}{lllll}
        \toprule
         & SHD & SSC & DVS Gesture &  \\
         & & & FC/CNN4 & VGG9 \\
         \midrule
         Architecture & [256$\times$3] & [512$\times$3] & \ref{appendix:dvsgesture_model_arch} & \ref{appendix:dvsgesture_model_arch}  \\
         Timestep size & 3.6 ms & 3.6 ms & 3 ms & 12 ms \\
         Number of frames & 250 & 250 & 2000 & 500 \\
         Batch size & 32 & 32 & 64 & 64 \\
         Epochs & 25 & 25 & 100 & 70 \\
         Initial learning rate & 1e-3* & 1e-3* & 1e-3 & 1e-3 \\
         Learning rate schedule & Step & Step & None & Cosine \\
         Schedule step size & 10 & 10 & &  \\
         Schedule step gamma & 0.1 & 0.1 & &  \\
         Schedule $\eta_\text{min}$ & & & & 1e-6 \\
         Schedule $T_\text{max}$ & & & & 70 \\
         Minimum threshold $\vartheta_0$ & 0.2 & 0.2 & 0.2 & 0.3 \\
         Number of buckets $K$ & 10 & 7 & 10 & 10 \\
         Bucket transfer rate $F$ & 0.15 & 0.15 & 0.05 & 0.2 \\
         Dropout & 0.1 & 0.1 & 0 & 0 \\
         Loss function & CE & CE warm-up (\ref{appendix:ce_warmup}) & CE & CE \\
         \bottomrule
    \end{tabular}
    \label{tab:hyperparameters}
\end{table}

As optimizer, for experiments with SHD, SSC, and DVS Gesture, Adam is used with $\beta=(0.9,0.999)$. For the delay and coincidence detection tasks, SGD is used. Hyperparameters were tuned by manual trial and error without a strict tuning budget or predefined search ranges.

\subsubsection{DVS Gesture model architecture}\label{appendix:dvsgesture_model_arch}
\begin{figure}[H]
    \centering
    \begin{subfigure}[t]{0.3\linewidth}
        \centering
        \includegraphics[width=\linewidth]{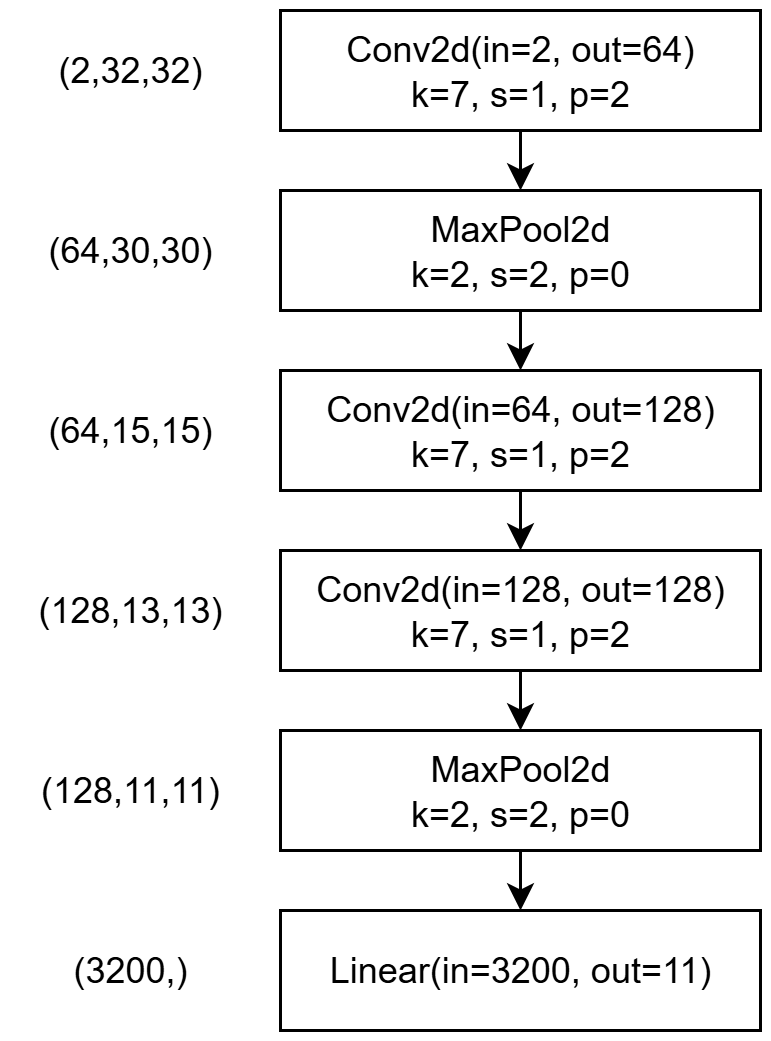}
        \caption{4-layer CNN architecture}
        \label{fig:cnn_dvsgesture_architecture}
    \end{subfigure}
    \hfill
    \begin{subfigure}[t]{0.6\linewidth}
        \centering
        \includegraphics[width=\linewidth]{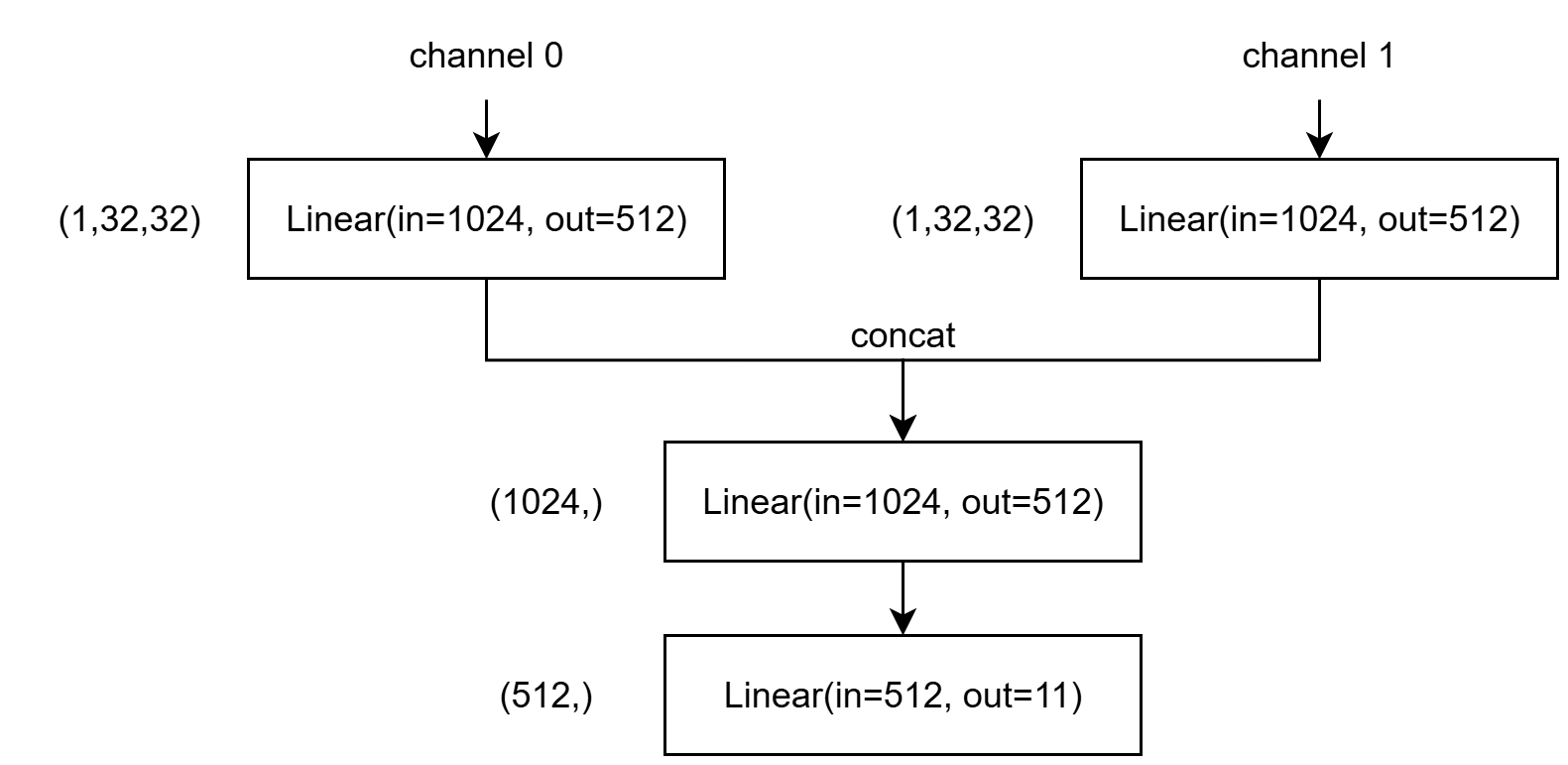}
        \caption{Fully-connected architecture}
        \label{fig:fc_dvsgesture_architecture}
    \end{subfigure}
    \hfill
    \begin{subfigure}[t]{0.8\linewidth}
        \centering
        \includegraphics[width=\linewidth]{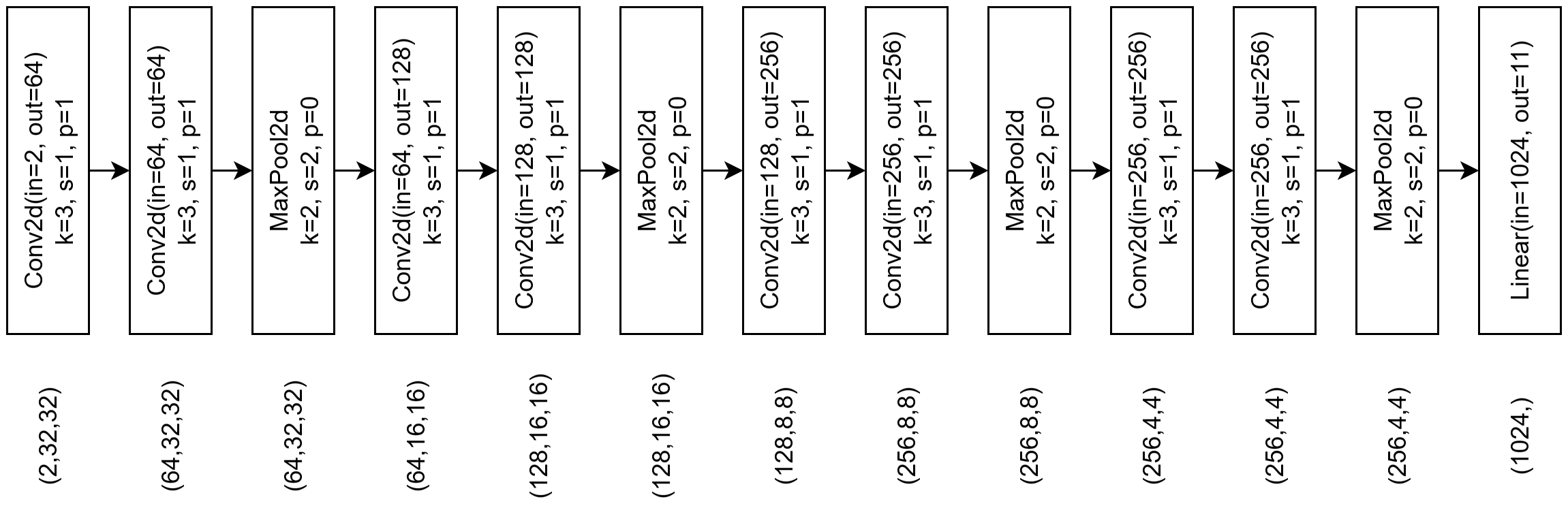}
        \caption{VGG9 architecture}
        \label{fig:vgg9_dvsgesture_architecture}
    \end{subfigure}
    
    \caption{(a) shows the 4-layer CNN architecture used for DVS Gesture (similar to what was used for DECOLLE in \cite{kaiser2020synaptic}).
    (b) shows the fully-connected architecture (similar to what was used for FPTT in \cite{yin2023fptt}).
    (c) shows the VGG9 architecture used for spatial scaling. Only for VGG9, the feature extractor (i.e., all layers before the classifier) uses weight standardization (see Section \ref{appendix:weight_standardization}) and group normalization (see Section \ref{appendix:group_norm}) instead of layer normalization.
    In all figures, neuron layers (including normalization and dropout, if applicable) are omitted; they are placed after each Conv2d or Linear layer.}
    \label{fig:dvsgesture_architectures}
\end{figure}

\subsection{Performance metrics}\label{appendix:metrics}

\subsubsection{CE warm-up loss}\label{appendix:ce_warmup}
For SSC, there is a delay in sample onset. This can be problematic when training online. To improve performance, we introduce a custom loss function that trains to mute the output at the start of the sample. It interpolates between MSE to zero and CE loss using a time-dependent weight:
\begin{equation}
L(t) = \beta(t) \cdot L^\text{MSE}(\hat{y}_\text{out}(t), 0) + (1 - \beta(t)) \cdot L^\text{CE}(y_\text{out}(t), y_\text{true}(t))
\end{equation}
where
\begin{equation}
\beta(t) = \beta_{\text{0}} \cdot (\beta_{\text{decay}})^{t}.
\end{equation}
with $\beta_{\text{decay}} = 0.99$ and $\beta_0=1.0$.

\subsubsection{Spike density}
\begin{equation}\label{eq:spike_density}
    \text{spike density} = \frac{1}{N_\text{samples} \cdot N_\text{neurons}} \sum_{i=1}^{N_\text{samples}} N_\text{spikes}[i]
\end{equation}
where $N_\text{samples}$ is the dataset size, $N_\text{neurons}$ is the total number of neurons in the hidden layers, and $N_\text{spikes}[i]$ is the spike count recorded during the inference of the $i$-th sample.

\subsection{Group normalization}\label{appendix:group_norm}
Group normalization divides the $C$ channels into $G$ groups (chosen as the largest value in $\{32,16,8,4,2,1\}$ such that $C \bmod G = 0$) and normalizes the neuron signal before rectification ($x$) based on group-wise statistics: $x_\text{norm} = \frac{x-\text{E}_g[x]}{\sqrt{\text{Var}_g[x]+\epsilon}}\cdot \gamma + \beta$, with $\gamma$ the gain, $\beta$ the bias (both trainable), and $\text{E}_g[x]$ and $\text{Var}_g[x]$ respectively the mean and variance of the neuron signal computed over all neurons within the same group.

\subsection{Weight standardization}\label{appendix:weight_standardization}
Weight standardization normalizes each output neuron's weights to zero mean and controlled variance. 
For $W \in \mathbb{R}^{d_{\text{out}} \times d_{\text{in}}}$, each row $W_i$ is transformed as
\begin{equation}
\hat{W}_i = \frac{W_i - \mathrm{E}[W_i]}{\sqrt{\mathrm{Var}[W_i]\cdot d_{\text{in}} + \epsilon}},
\end{equation}
where $\mathrm{E}[W_i]$ and $\mathrm{Var}[W_i]$ denote the mean and variance of the weights of neuron $i$. 

\subsection{Computer resources}
Our experiments were done on a workstation equipped with 256 GB of memory, an NVIDIA RTX A6000 GPU with 49 GB of VRAM, and an Intel Xeon Bronze 3206R CPU.

\subsection{Experimental setup delay experiment}\label{sec:delay_experiment_setup}
\begin{table}[H]
  \centering
  \caption{Hyperparameters for the delay experiment.}
  \label{tab:hyperparameters_delay_experiment}
  \begin{tabular}{@{}ll@{}}
    \toprule
    \textbf{Hyperparameter} & \textbf{Value} \\
    \midrule
    Threshold & 0.2 \\
    Batch size & 1 \\
    Epochs & Until converged \\
    Optimizer  & SGD \\
    Learning rate & $10^{-1}$ \\
    Synaptic weight (frozen) & 5.0 \\
    Synaptic weight bias (frozen) & 0.0 \\
    Bucket weight initialization & $\mathcal{N}(\mu = 0.1,\, \sigma = 0.01)$; fixed seed \\
    Number of buckets $K$ & 20 \\
    Bucket transfer rate $F$ & 0.3 \\
    Loss & MSE \\
    Normalization & None \\
    Adaptive threshold & Disabled \\
    Surrogate (BPTT only) & arctan ($\alpha = 2$) \\
    \bottomrule
  \end{tabular}
\end{table}

\newpage
\input{checklist.tex}

\end{document}

%% file: checklist.tex
\section*{NeurIPS Paper Checklist}

\begin{enumerate}

\item {\bf Claims}
    \item[] Question: Do the main claims made in the abstract and introduction accurately reflect the paper's contributions and scope?
    \item[] Answer: \answerYes{} % Replace by \answerYes{}, \answerNo{}, or \answerNA{}.
    \item[] Justification: Our results support our claims.
    \item[] Guidelines:
    \begin{itemize}
        \item The answer \answerNA{} means that the abstract and introduction do not include the claims made in the paper.
        \item The abstract and/or introduction should clearly state the claims made, including the contributions made in the paper and important assumptions and limitations. A \answerNo{} or \answerNA{} answer to this question will not be perceived well by the reviewers. 
        \item The claims made should match theoretical and experimental results, and reflect how much the results can be expected to generalize to other settings. 
        \item It is fine to include aspirational goals as motivation as long as it is clear that these goals are not attained by the paper. 
    \end{itemize}

\item {\bf Limitations}
    \item[] Question: Does the paper discuss the limitations of the work performed by the authors?
    \item[] Answer: \answerYes{} % Replace by \answerYes{}, \answerNo{}, or \answerNA{}.
    \item[] Justification: The limitations are discussed in the results and discussion sections.
    \item[] Guidelines:
    \begin{itemize}
        \item The answer \answerNA{} means that the paper has no limitation while the answer \answerNo{} means that the paper has limitations, but those are not discussed in the paper. 
        \item The authors are encouraged to create a separate ``Limitations'' section in their paper.
        \item The paper should point out any strong assumptions and how robust the results are to violations of these assumptions (e.g., independence assumptions, noiseless settings, model well-specification, asymptotic approximations only holding locally). The authors should reflect on how these assumptions might be violated in practice and what the implications would be.
        \item The authors should reflect on the scope of the claims made, e.g., if the approach was only tested on a few datasets or with a few runs. In general, empirical results often depend on implicit assumptions, which should be articulated.
        \item The authors should reflect on the factors that influence the performance of the approach. For example, a facial recognition algorithm may perform poorly when image resolution is low or images are taken in low lighting. Or a speech-to-text system might not be used reliably to provide closed captions for online lectures because it fails to handle technical jargon.
        \item The authors should discuss the computational efficiency of the proposed algorithms and how they scale with dataset size.
        \item If applicable, the authors should discuss possible limitations of their approach to address problems of privacy and fairness.
        \item While the authors might fear that complete honesty about limitations might be used by reviewers as grounds for rejection, a worse outcome might be that reviewers discover limitations that aren't acknowledged in the paper. The authors should use their best judgment and recognize that individual actions in favor of transparency play an important role in developing norms that preserve the integrity of the community. Reviewers will be specifically instructed to not penalize honesty concerning limitations.
    \end{itemize}

\item {\bf Theory assumptions and proofs}
    \item[] Question: For each theoretical result, does the paper provide the full set of assumptions and a complete (and correct) proof?
    \item[] Answer: \answerNA{} % Replace by \answerYes{}, \answerNo{}, or \answerNA{}.
    \item[] Justification: No theoretical results.
    \item[] Guidelines:
    \begin{itemize}
        \item The answer \answerNA{} means that the paper does not include theoretical results. 
        \item All the theorems, formulas, and proofs in the paper should be numbered and cross-referenced.
        \item All assumptions should be clearly stated or referenced in the statement of any theorems.
        \item The proofs can either appear in the main paper or the supplemental material, but if they appear in the supplemental material, the authors are encouraged to provide a short proof sketch to provide intuition. 
        \item Inversely, any informal proof provided in the core of the paper should be complemented by formal proofs provided in appendix or supplemental material.
        \item Theorems and Lemmas that the proof relies upon should be properly referenced. 
    \end{itemize}

    \item {\bf Experimental result reproducibility}
    \item[] Question: Does the paper fully disclose all the information needed to reproduce the main experimental results of the paper to the extent that it affects the main claims and/or conclusions of the paper (regardless of whether the code and data are provided or not)?
    \item[] Answer: \answerYes{} % Replace by \answerYes{}, \answerNo{}, or \answerNA{}.
    \item[] Justification: Besides the methods that describe the general idea of our approach, we included a detailed overview of the hyperparameters in the appendix. Furthermore, we included the same code as we used for our experiments. This should enable full reproducibility.
    \item[] Guidelines:
    \begin{itemize}
        \item The answer \answerNA{} means that the paper does not include experiments.
        \item If the paper includes experiments, a \answerNo{} answer to this question will not be perceived well by the reviewers: Making the paper reproducible is important, regardless of whether the code and data are provided or not.
        \item If the contribution is a dataset and\slash or model, the authors should describe the steps taken to make their results reproducible or verifiable. 
        \item Depending on the contribution, reproducibility can be accomplished in various ways. For example, if the contribution is a novel architecture, describing the architecture fully might suffice, or if the contribution is a specific model and empirical evaluation, it may be necessary to either make it possible for others to replicate the model with the same dataset, or provide access to the model. In general. releasing code and data is often one good way to accomplish this, but reproducibility can also be provided via detailed instructions for how to replicate the results, access to a hosted model (e.g., in the case of a large language model), releasing of a model checkpoint, or other means that are appropriate to the research performed.
        \item While NeurIPS does not require releasing code, the conference does require all submissions to provide some reasonable avenue for reproducibility, which may depend on the nature of the contribution. For example
        \begin{enumerate}
            \item If the contribution is primarily a new algorithm, the paper should make it clear how to reproduce that algorithm.
            \item If the contribution is primarily a new model architecture, the paper should describe the architecture clearly and fully.
            \item If the contribution is a new model (e.g., a large language model), then there should either be a way to access this model for reproducing the results or a way to reproduce the model (e.g., with an open-source dataset or instructions for how to construct the dataset).
            \item We recognize that reproducibility may be tricky in some cases, in which case authors are welcome to describe the particular way they provide for reproducibility. In the case of closed-source models, it may be that access to the model is limited in some way (e.g., to registered users), but it should be possible for other researchers to have some path to reproducing or verifying the results.
        \end{enumerate}
    \end{itemize}

\item {\bf Open access to data and code}
    \item[] Question: Does the paper provide open access to the data and code, with sufficient instructions to faithfully reproduce the main experimental results, as described in supplemental material?
    \item[] Answer: \answerYes{} % Replace by \answerYes{}, \answerNo{}, or \answerNA{}.
    \item[] Justification: At submission time we provide anonymized code for reproducibility. After acceptance we will release a package that enables anyone to easily get started with SpikingGamma.
    \item[] Guidelines:
    \begin{itemize}
        \item The answer \answerNA{} means that paper does not include experiments requiring code.
        \item Please see the NeurIPS code and data submission guidelines (\url{https://neurips.cc/public/guides/CodeSubmissionPolicy}) for more details.
        \item While we encourage the release of code and data, we understand that this might not be possible, so \answerNo{} is an acceptable answer. Papers cannot be rejected simply for not including code, unless this is central to the contribution (e.g., for a new open-source benchmark).
        \item The instructions should contain the exact command and environment needed to run to reproduce the results. See the NeurIPS code and data submission guidelines (\url{https://neurips.cc/public/guides/CodeSubmissionPolicy}) for more details.
        \item The authors should provide instructions on data access and preparation, including how to access the raw data, preprocessed data, intermediate data, and generated data, etc.
        \item The authors should provide scripts to reproduce all experimental results for the new proposed method and baselines. If only a subset of experiments are reproducible, they should state which ones are omitted from the script and why.
        \item At submission time, to preserve anonymity, the authors should release anonymized versions (if applicable).
        \item Providing as much information as possible in supplemental material (appended to the paper) is recommended, but including URLs to data and code is permitted.
    \end{itemize}

\item {\bf Experimental setting/details}
    \item[] Question: Does the paper specify all the training and test details (e.g., data splits, hyperparameters, how they were chosen, type of optimizer) necessary to understand the results?
    \item[] Answer: \answerYes{} % Replace by \answerYes{}, \answerNo{}, or \answerNA{}.
    \item[] Justification: Full details are provided in the main text and appendix.
    \item[] Guidelines:
    \begin{itemize}
        \item The answer \answerNA{} means that the paper does not include experiments.
        \item The experimental setting should be presented in the core of the paper to a level of detail that is necessary to appreciate the results and make sense of them.
        \item The full details can be provided either with the code, in appendix, or as supplemental material.
    \end{itemize}

\item {\bf Experiment statistical significance}
    \item[] Question: Does the paper report error bars suitably and correctly defined or other appropriate information about the statistical significance of the experiments?
    \item[] Answer: \answerYes{} % Replace by \answerYes{}, \answerNo{}, or \answerNA{}.
    \item[] Justification: All experiments were repeated multiple times and errors bars / standard deviations were included.
    \item[] Guidelines:
    \begin{itemize}
        \item The answer \answerNA{} means that the paper does not include experiments.
        \item The authors should answer \answerYes{} if the results are accompanied by error bars, confidence intervals, or statistical significance tests, at least for the experiments that support the main claims of the paper.
        \item The factors of variability that the error bars are capturing should be clearly stated (for example, train/test split, initialization, random drawing of some parameter, or overall run with given experimental conditions).
        \item The method for calculating the error bars should be explained (closed form formula, call to a library function, bootstrap, etc.)
        \item The assumptions made should be given (e.g., Normally distributed errors).
        \item It should be clear whether the error bar is the standard deviation or the standard error of the mean.
        \item It is OK to report 1-sigma error bars, but one should state it. The authors should preferably report a 2-sigma error bar than state that they have a 96\% CI, if the hypothesis of Normality of errors is not verified.
        \item For asymmetric distributions, the authors should be careful not to show in tables or figures symmetric error bars that would yield results that are out of range (e.g., negative error rates).
        \item If error bars are reported in tables or plots, the authors should explain in the text how they were calculated and reference the corresponding figures or tables in the text.
    \end{itemize}

\item {\bf Experiments compute resources}
    \item[] Question: For each experiment, does the paper provide sufficient information on the computer resources (type of compute workers, memory, time of execution) needed to reproduce the experiments?
    \item[] Answer: \answerYes{} % Replace by \answerYes{}, \answerNo{}, or \answerNA{}.
    \item[] Justification: Computer resources used are described in the appendix. These can cover all of the experiments.
    \item[] Guidelines:
    \begin{itemize}
        \item The answer \answerNA{} means that the paper does not include experiments.
        \item The paper should indicate the type of compute workers CPU or GPU, internal cluster, or cloud provider, including relevant memory and storage.
        \item The paper should provide the amount of compute required for each of the individual experimental runs as well as estimate the total compute. 
        \item The paper should disclose whether the full research project required more compute than the experiments reported in the paper (e.g., preliminary or failed experiments that didn't make it into the paper). 
    \end{itemize}
    
\item {\bf Code of ethics}
    \item[] Question: Does the research conducted in the paper conform, in every respect, with the NeurIPS Code of Ethics \url{https://neurips.cc/public/EthicsGuidelines}?
    \item[] Answer: \answerYes{} % Replace by \answerYes{}, \answerNo{}, or \answerNA{}.
    \item[] Justification: No remarks.
    \item[] Guidelines:
    \begin{itemize}
        \item The answer \answerNA{} means that the authors have not reviewed the NeurIPS Code of Ethics.
        \item If the authors answer \answerNo, they should explain the special circumstances that require a deviation from the Code of Ethics.
        \item The authors should make sure to preserve anonymity (e.g., if there is a special consideration due to laws or regulations in their jurisdiction).
    \end{itemize}

\item {\bf Broader impacts}
    \item[] Question: Does the paper discuss both potential positive societal impacts and negative societal impacts of the work performed?
    \item[] Answer: \answerYes{} % Replace by \answerYes{}, \answerNo{}, or \answerNA{}.
    \item[] Justification: Potential impact is discussed in the introduction.
    \item[] Guidelines:
    \begin{itemize}
        \item The answer \answerNA{} means that there is no societal impact of the work performed.
        \item If the authors answer \answerNA{} or \answerNo, they should explain why their work has no societal impact or why the paper does not address societal impact.
        \item Examples of negative societal impacts include potential malicious or unintended uses (e.g., disinformation, generating fake profiles, surveillance), fairness considerations (e.g., deployment of technologies that could make decisions that unfairly impact specific groups), privacy considerations, and security considerations.
        \item The conference expects that many papers will be foundational research and not tied to particular applications, let alone deployments. However, if there is a direct path to any negative applications, the authors should point it out. For example, it is legitimate to point out that an improvement in the quality of generative models could be used to generate Deepfakes for disinformation. On the other hand, it is not needed to point out that a generic algorithm for optimizing neural networks could enable people to train models that generate Deepfakes faster.
        \item The authors should consider possible harms that could arise when the technology is being used as intended and functioning correctly, harms that could arise when the technology is being used as intended but gives incorrect results, and harms following from (intentional or unintentional) misuse of the technology.
        \item If there are negative societal impacts, the authors could also discuss possible mitigation strategies (e.g., gated release of models, providing defenses in addition to attacks, mechanisms for monitoring misuse, mechanisms to monitor how a system learns from feedback over time, improving the efficiency and accessibility of ML).
    \end{itemize}
    
\item {\bf Safeguards}
    \item[] Question: Does the paper describe safeguards that have been put in place for responsible release of data or models that have a high risk for misuse (e.g., pre-trained language models, image generators, or scraped datasets)?
    \item[] Answer: \answerNA{} % Replace by \answerYes{}, \answerNo{}, or \answerNA{}.
    \item[] Justification: Not high risk.
    \item[] Guidelines:
    \begin{itemize}
        \item The answer \answerNA{} means that the paper poses no such risks.
        \item Released models that have a high risk for misuse or dual-use should be released with necessary safeguards to allow for controlled use of the model, for example by requiring that users adhere to usage guidelines or restrictions to access the model or implementing safety filters. 
        \item Datasets that have been scraped from the Internet could pose safety risks. The authors should describe how they avoided releasing unsafe images.
        \item We recognize that providing effective safeguards is challenging, and many papers do not require this, but we encourage authors to take this into account and make a best faith effort.
    \end{itemize}

\item {\bf Licenses for existing assets}
    \item[] Question: Are the creators or original owners of assets (e.g., code, data, models), used in the paper, properly credited and are the license and terms of use explicitly mentioned and properly respected?
    \item[] Answer: \answerYes{} % Replace by \answerYes{}, \answerNo{}, or \answerNA{}.
    \item[] Justification: Done.
    \item[] Guidelines:
    \begin{itemize}
        \item The answer \answerNA{} means that the paper does not use existing assets.
        \item The authors should cite the original paper that produced the code package or dataset.
        \item The authors should state which version of the asset is used and, if possible, include a URL.
        \item The name of the license (e.g., CC-BY 4.0) should be included for each asset.
        \item For scraped data from a particular source (e.g., website), the copyright and terms of service of that source should be provided.
        \item If assets are released, the license, copyright information, and terms of use in the package should be provided. For popular datasets, \url{paperswithcode.com/datasets} has curated licenses for some datasets. Their licensing guide can help determine the license of a dataset.
        \item For existing datasets that are re-packaged, both the original license and the license of the derived asset (if it has changed) should be provided.
        \item If this information is not available online, the authors are encouraged to reach out to the asset's creators.
    \end{itemize}

\item {\bf New assets}
    \item[] Question: Are new assets introduced in the paper well documented and is the documentation provided alongside the assets?
    \item[] Answer: \answerYes{} % Replace by \answerYes{}, \answerNo{}, or \answerNA{}.
    \item[] Justification: Code includes documentation.
    \item[] Guidelines:
    \begin{itemize}
        \item The answer \answerNA{} means that the paper does not release new assets.
        \item Researchers should communicate the details of the dataset\slash code\slash model as part of their submissions via structured templates. This includes details about training, license, limitations, etc. 
        \item The paper should discuss whether and how consent was obtained from people whose asset is used.
        \item At submission time, remember to anonymize your assets (if applicable). You can either create an anonymized URL or include an anonymized zip file.
    \end{itemize}

\item {\bf Crowdsourcing and research with human subjects}
    \item[] Question: For crowdsourcing experiments and research with human subjects, does the paper include the full text of instructions given to participants and screenshots, if applicable, as well as details about compensation (if any)? 
    \item[] Answer: \answerNA{} % Replace by \answerYes{}, \answerNo{}, or \answerNA{}.
    \item[] Justification: No crowdsourcing.
    \item[] Guidelines:
    \begin{itemize}
        \item The answer \answerNA{} means that the paper does not involve crowdsourcing nor research with human subjects.
        \item Including this information in the supplemental material is fine, but if the main contribution of the paper involves human subjects, then as much detail as possible should be included in the main paper. 
        \item According to the NeurIPS Code of Ethics, workers involved in data collection, curation, or other labor should be paid at least the minimum wage in the country of the data collector. 
    \end{itemize}

\item {\bf Institutional review board (IRB) approvals or equivalent for research with human subjects}
    \item[] Question: Does the paper describe potential risks incurred by study participants, whether such risks were disclosed to the subjects, and whether Institutional Review Board (IRB) approvals (or an equivalent approval/review based on the requirements of your country or institution) were obtained?
    \item[] Answer: \answerNA{} % Replace by \answerYes{}, \answerNo{}, or \answerNA{}.
    \item[] Justification: Paper does not involve crowdsourcing nor research with human subjects.
    \item[] Guidelines:
    \begin{itemize}
        \item The answer \answerNA{} means that the paper does not involve crowdsourcing nor research with human subjects.
        \item Depending on the country in which research is conducted, IRB approval (or equivalent) may be required for any human subjects research. If you obtained IRB approval, you should clearly state this in the paper. 
        \item We recognize that the procedures for this may vary significantly between institutions and locations, and we expect authors to adhere to the NeurIPS Code of Ethics and the guidelines for their institution. 
        \item For initial submissions, do not include any information that would break anonymity (if applicable), such as the institution conducting the review.
    \end{itemize}

\item {\bf Declaration of LLM usage}
    \item[] Question: Does the paper describe the usage of LLMs if it is an important, original, or non-standard component of the core methods in this research? Note that if the LLM is used only for writing, editing, or formatting purposes and does \emph{not} impact the core methodology, scientific rigor, or originality of the research, declaration is not required.
    %this research? 
    \item[] Answer: \answerNA{} % Replace by \answerYes{}, \answerNo{}, or \answerNA{}.
    \item[] Justification: Core method development in this research does not involve LLMs as any important, original, or non-standard components.
    \item[] Guidelines:
    \begin{itemize}
        \item The answer \answerNA{} means that the core method development in this research does not involve LLMs as any important, original, or non-standard components.
        \item Please refer to our LLM policy in the NeurIPS handbook for what should or should not be described.
    \end{itemize}

\end{enumerate}